\documentclass{article}

\usepackage[preprint]{neurips_2026}

\usepackage[utf8]{inputenc}
\usepackage[T1]{fontenc}
\usepackage{microtype}
\usepackage{amsfonts}
\usepackage{booktabs}
\usepackage{nicefrac}
\usepackage{url}
\usepackage{hyperref}
\usepackage{xcolor}
\usepackage{graphicx}
\usepackage{caption}
\usepackage{float}
\usepackage{placeins}
\usepackage{tabularx}
\usepackage{array}
\usepackage{chngcntr}

\title{EXPERT-VALIDATED STEM QA}

\author{%
Kihwan~Han, \quad Saurabh~Patil, \quad Chinmayee~Shukla, \quad Abhinav~Sharma, \\
\textbf{Marko~Pavlovic}$^{*}$, \quad \textbf{Anshuman~Lall}$^{*}$, \quad \textbf{Mahesh~Joshi} \\
Turing\\
${}^{*}$Work done while at Turing\\
\texttt{kihwan.han@turing.com}.
}

\begin{document}
\raggedbottom

\maketitle

\begin{abstract}

Recent advancements in AI are helping scientists achieve breakthroughs in fields such as mathematics, medicine, and materials sciences. New evaluation datasets for AI models contribute to such advancement in AI. In the STEM domain, frontier models have consumed most of the available online data, creating the need for human-created datasets that codify the knowledge of leading experts in the domain. There are several STEM datasets available for the research community in this field. However, there are some gaps in these datasets, leaving room for improvement. Examples of gaps include (1) saturation in model performance on these datasets, leaving no head-room for meaningful evaluations, (2) skewed taxonomy distributions, (3) multiple choice question format that is misaligned with how scientists use AI in the real world, and (4) inaccurate answers and rationales partially led by a contest-based data collection and a time-bound review process. In this study, we present ‘Expert-validated STEM QA’, a high-quality, expert-validated STEM dataset (N=398) in Physics, Chemistry, Biology, and Mathematics, created by 241 domain experts. We (1) carefully designed a taxonomy with balanced distribution, (2) vetted question contributors with quality-driven incentive, (3) conducted multiple rounds of reviews with revisions validated by domain experts based on consensus, and (4) created the dataset in verifiable question and answer format. Our study demonstrated low performance (<25\%) of frontier AI models on the dataset as a benchmark. Post-training on a separate, private version of the dataset (N=2,000) increased performance of the open source model by 15\% relative to the baseline model (p=0.045) on the STEM subset of HLE-verified dataset, indicating potential utility of the dataset for model training. We have open-sourced a portion of our dataset for the AI research community at \url{https://huggingface.co/datasets/TuringEnterprises/Open-RL}.

\end{abstract}

\section{INTRODUCTION}

Science advances by building on prior knowledge — either refining or refuting it. This incremental nature demands multi-step reasoning: scientists must divide inquiries into subtasks, formulating hierarchical hypotheses, and navigating several options to make a conclusion. Recently, rapid progress of artificial intelligence (AI) is transforming numerous fields, including scientific research. For example, AI has enabled biologists to predict protein structures and biomolecular complex structures, including protein–ligand and protein–nucleic-acid complexes, addressing long-standing challenges in structural biology and drug discovery \cite{abramson_accurate_2024}. Further, advancement in reasoning models (\cite{li_system_2025} for a review) has increased scientists' adoption of AI for research. Accordingly, there are multiple datasets to measure capability of AI for scientific reasoning \cite{rein_gpqa_2023, phan_humanitys_2025, zhai_hle-verified_2026, team_supergpqa_2025, lu_scp-116k_2025, ma_sci-reason_2025, moshkov_aimo-2_2025, glazer_frontiermath_2025, noauthor_nvidiaopensciencereasoning-2_2026, noauthor_futurehouseether0-benchmark_2025}.

However, existing benchmarks exhibit notable gaps. For example, some benchmarks are already near saturation for state-of-the-art models \cite{rein_gpqa_2023}. In some cases, the reliance on multiple-choice or fill-in-the-blank formats \cite{rein_gpqa_2023, team_supergpqa_2025, phan_humanitys_2025, zhai_hle-verified_2026} misaligns with how scientists typically use AI in practice. Some datasets exhibit skewed taxonomy distributions \cite{phan_humanitys_2025, zhai_hle-verified_2026} or restricted to certain domains \cite{moshkov_aimo-2_2025, glazer_frontiermath_2025, noauthor_futurehouseether0-benchmark_2025}. Several datasets were curated from online resources \cite{team_supergpqa_2025, lu_scp-116k_2025, ma_sci-reason_2025}, raising concerns about data leakage into model training. Most importantly, datasets may suffer from inaccurate answers or ambiguous questions \cite{phan_humanitys_2025}, presumably owing to contest-based data collection mechanisms and a time-bound review process. Some datasets lack human verification entirely \cite{lu_scp-116k_2025, moshkov_aimo-2_2025}. In particular, inaccurate answers without thorough expert reviews would be critical issues to scientists as potential users of AI because model performance scores derived from flawed benchmarks misrepresent true capability. Thus, it can potentially inflate a model's apparent confidence when responding to scientists' questions, which in turn negatively impacts the scientific community.


Here, we present ‘Expert-validated STEM QA’, a high-quality and expert-validated STEM dataset (N=398) spanning Physics, Chemistry, Biology, and Mathematics (PCMB; see Fig. 1 for examples). The dataset was constructed using a carefully designed taxonomy, quality-driven incentive structures, and consensus-based multi-layer expert review. In this study, we describe the data generation workflow, demonstrate the dataset's utility for benchmarking model performance, and present a case study on using the dataset to improve scientific reasoning in existing models.


\begin{figure}[htbp]
\centering
\includegraphics[width=0.90\linewidth]{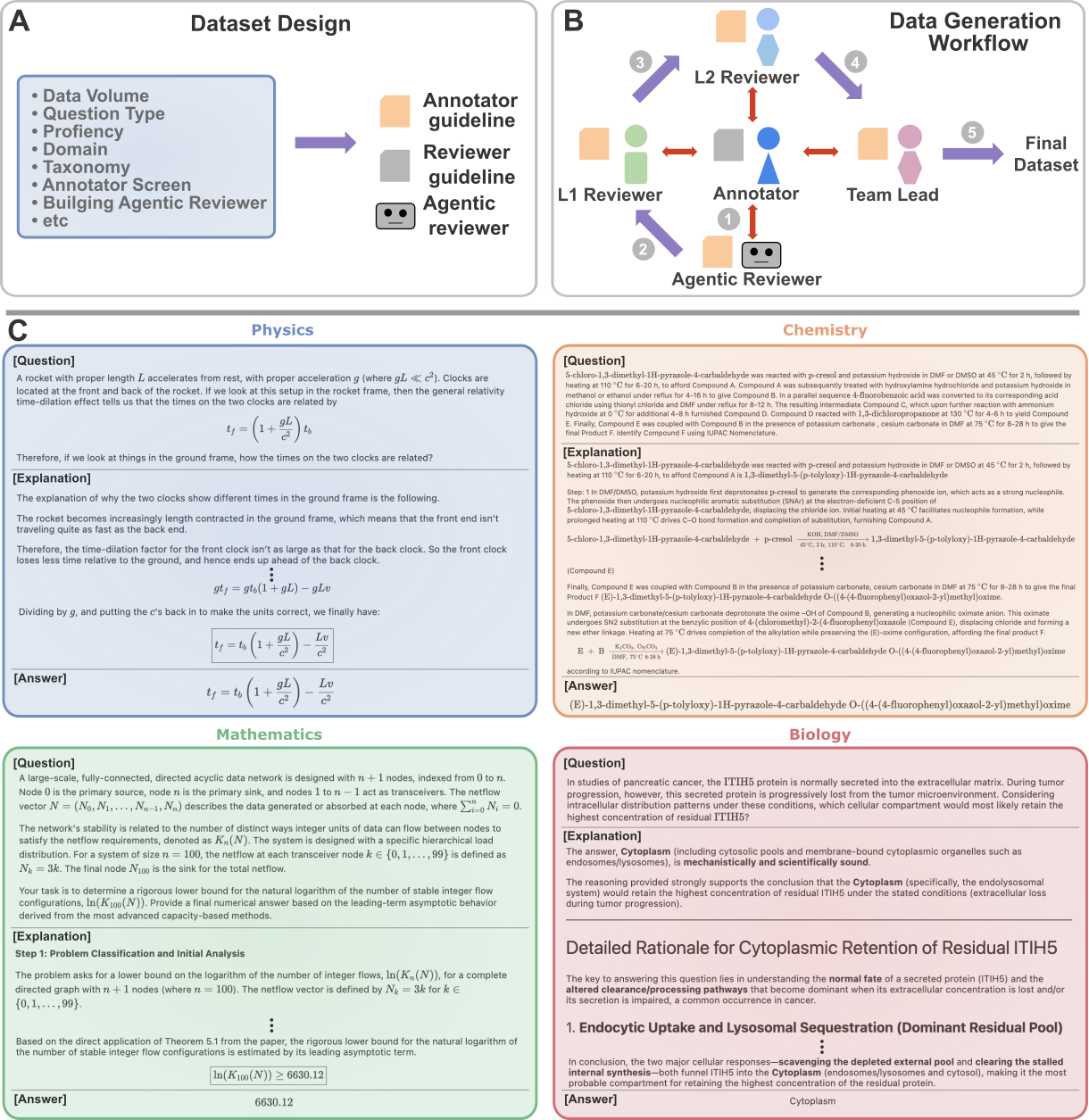}
\caption{Dataset generation process. A: We designed the dataset with specifications on data volume, domain, taxonomy examples, etc., yielding annotation and review guidelines with agentic reviewers. B: Based on the artifacts, annotators generated the data under four-levels of review from agentic reviewer to team leads with revisions throughout the workflow. C: Examples of the dataset from Physics, Chemistry, Mathematics, and Biology, respectively. The dataset consists of a (question, explanation, answer) triplet.}
\label{fig:process_examples}
\end{figure}


\section{MATERIALS AND METHODS}

\subsection{Dataset generation process}

Overall, the dataset generation process comprises dataset design, annotation, and reviews (Fig. 1).


\subsubsection{Dataset design}

First, we consulted subject matter experts (SMEs) and team leads to define the attributes of a prospective dataset considering current gaps in scientific AI benchmarks and practical constraints on data generation. We then specified the dataset design (Fig. 1A), defining: (1) (1) the structure of a single data instance (question, explanation, and answer), (2) question format (verifiable question and answer), (3) level of proficiency, (4) data volume, (5) key qualifications for annotators and reviewers, (6) domains (PCMB), (7) taxonomy that ensures topical diversity, (8) review criteria to ensure data quality - including non-searchability, factual accuracy, and multi-step reasoning requirements, and (9) which review criteria to automate via agentic reviewers. This process produced three artifacts: an annotator guideline, a reviewer guideline, and agentic reviewers.


\paragraph{Annotator guideline}

To standardize examples across the dataset, we provided annotators with clear guidelines to follow. The key items in the guideline include project goal, definition of a single instance, taxonomy, metadata, and difficulty level calibrated against responses from mid-tier models.


Taxonomy comprises two levels (L1 and L2) under each domain. To reflect the balance between historical foundations, conceptual breadth, and modern research development in the taxonomy, we combined the following approaches: (1) extension of existing taxonomy from GPQA \cite{rein_gpqa_2023}, (2) reference to research areas from prestigious scientific journals in each of the subjects, (3) review of key research papers in the subject covering theory, methods, and applications, and (4) discussion among SMEs in the team based on their experiences and expertise. Rather than requiring uniform coverage across taxonomy categories, we instructed annotators to contribute questions within their own areas of expertise (refer to Appendix A for full taxonomy details).


To facilitate expert review, annotators were required to document metadata for each instance. The metadata includes evidence that the question is not directly answerable via web search, and model response evaluations confirming that any failures stem from reasoning limitations rather than question ambiguity.


\paragraph{Reviewer guideline}

The reviewer guideline is the primary reference document by which reviewers assess annotator-generated examples. It specifies the review rubrics in detail (Table S1), covering all required data quality criteria. Recognizing that answer accuracy is a critical determinant of benchmark reliability \cite{zhai_hle-verified_2026}, we designed a structured, multi-stage review process. Further, to maximize inter-rater agreement among domain experts, each rubric item was decomposed into atomic, binary judgments (e.g., Pass/Fail), rather than graded scales (e.g., 1–5 Likert scale) \cite{mallinar_scalable_2025, botelho_scale_2025}.


\paragraph{Review agents}

A key strategic objective was to optimize human effort; accordingly, we identified the subset of rubric items for which agentic reviewers could reliably assess data quality, and built automated review agents for those items. We first identified highly trusted reviewers from each domain, who then constructed a golden review set from small samples, held out from the dataset. From the golden review dataset, we trained and calibrated the review agents using standard train/validation/test splits until both precision and recall on the held-out test split reached a minimum threshold of 80\%. This integration freed human reviewers to apply their expertise to the more cognitively demanding rubric items requiring nuanced judgment.


\subsubsection{Talent pool of annotators, reviewers, and team leads}

Annotator recruitment proceeded in two stages: a multi-phase screening process followed by structured training. To attract high-caliber contributors, we employed a rigorous screening process emphasizing cognitive agility and subject matter expertise. We sourced specialists from elite academic institutions to initially vet candidates via their research profiles and scholarly records. The selection process involved two distinct phases: an initial \textit{domain-agnostic reasoning assessment} to assess general reasoning ability, followed by a \textit{specialized domain challenge}. In the latter, candidates formulated challenging technical problems and accompanying evaluation rubrics. These submissions were audited by an \textit{agentic reviewer} to verify logical density and structural clarity. Final technical interviews were led by senior team leads. Admitted annotators then completed intensive bootcamp training on producing (question, explanation, answer) triplets. To ensure dataset quality, approximately 90\% of participants were off-boarded at post-bootcamp. The remaining contributors - primarily doctoral researchers with 5–10 years of domain experience - were nominated as reviewers. These reviewers received supplementary training to serve as expert validators. Team leads were senior members with extensive academic or industry experience; a substantial proportion held faculty or tenure-track positions. Team leads conducted regular meetings with annotators and reviewers to maintain data quality throughout the process.


\subsubsection{Annotation and review process}

Annotators constructed the dataset in accordance with the annotator guidelines and worked closely with the agentic reviewers, reviewers, and team leads throughout multiple rounds of revisions (Fig. 1B). The annotators first submitted the examples to the agentic reviewers.The agentic reviewers assessed lower-level data quality criteria, including linguistic clarity and web-searchability (Table S1). Annotators either revised their examples accordingly or formally disputed the review outcome with documented justification. L1 human reviewers then evaluated each example for surface-level integrity and formatting consistency. This encompassed verifying format requirements, checking completeness, and confirming compliance with the project style guide (Table S1). To maximize factual accuracy and mitigate individual reviewer bias, the L2 human reviewers focused on the scientific validity and factual accuracy of the (question, explanation, answer) triplets. This two-tier human review process produced a robust, expert-verified dataset. Lastly, team leads spot-checked the final version of each instance. Representative examples from each domain are shown in Fig. 1C.


\subsection{Dataset analysis}

We specified the generated dataset in multiple dimensions, followed by assessment of model performance on the dataset.

\subsubsection{Dataset specifications}

We identified the total volume of the dataset, domain and taxonomy breakdowns of the dataset, and distributions of token counts for question and combined response (explanation and answer).

\subsubsection{Assessment of model performance on the dataset}

We evaluated a total of 10 model variants spanning three proprietary and one open-source model family, each configured with varying reasoning effort levels, and assessed on our dataset. The proprietary models included Claude Opus 4.5 (claude-opus-4-5-20251101) Extended Thinking with low and high reasoning efforts \cite{anthropic_claude_2025}, Gemini 3.1 Pro (gemini-3.1-pro-preview) with low and high thinking levels \cite{google_gemini_2026}, GPT 5.2 Pro (gpt-5.2-pro-2025-12-11) with medium and high reasoning efforts \cite{openai_chatgpt_2025}. The open source model included a quantized version (Q5\_K\_M; recommended option per Unsloth \cite{daniel_han_unsloth_2023}) of Qwen 3.5 27B and 9B with reasoning enabled and disabled \cite{qwen_team_qwen35_2026}, sourced from Unsloth. We selected these smaller Qwen 3.5 variants because they are cost- and time-efficient to run the evaluations and subsequent post-training experiments. Prior evaluations demonstrated that these smaller variants perform competitively relative to larger Qwen 3.5 models on the existing benchmarks in STEM reasoning \cite{noauthor_unslothqwen35-27b-gguf_2026, noauthor_unslothqwen35-9b-gguf_2026}.


For all the evaluated models, we used a modified version of HLE prompts \cite{phan_humanitys_2025} for generating and grading responses (Figs. S1, S2). The responses include explanation, answer, and confidence. For response generation, across all configurations, the maximum completion token limit (encompassing reasoning trace and response) was set to 32,768 with a separate response token limit of 2,048 where applicable. The proprietary models generated responses with their default sampling parameters. The sampling parameters for the open source models were temperature=0, top p=0.95, top k=20, min p=0.0, and repeat penalty=1.0, referenced from \cite{noauthor_qwen35_2026}, with the exception of temperature, which was set to 0 to ensure deterministic outputs. Responses were graded using GPT-4o and Qwen 3.5-4B-UD-Q4\_K\_XL, validated against manual grading on a held-out subset to confirm agreement. For open-source models, response generation and grading were performed using llama.cpp on an RTX PRO 6000 GPU.


To assess response consistency, we computed pass rate and pass@k for the open-source reasoning models across 8 independent trials.


\subsection{Post-training}

To determine whether post-training on the dataset improves model accuracy, we conducted supervised fine-tuning (SFT) on Qwen 9B (reasoning disabled; BF16 precision) with LoRA (rank=16, alpha=32) \cite{hu_lora_2021} using unsloth \cite{daniel_han_unsloth_2023}. As we did not have reasoning traces to create labels for SFT, we disabled reasoning and used final answers alone as training labels. To accommodate this constraint, we revised the response prompt (Fig. S3) and label generation procedure (Fig. S4).


For SFT, we used a separate, proprietary dataset (N=2,000) partitioned into training and validation sets of 1,700 and 300 instances, respectively. Based on the token length distributions of questions and responses in this dataset, we set the maximum context window to 2,560 tokens and the maximum completion length to 512 tokens for fine-tuning. After hyperparameter search over learning rates of 2e-4, 2e-5, 5e-6, and 2e-6, we set up the following parameters for SFT: learning rate=2e-6, epochs=3, warm up=3, per device batch size=1, gradient accumulation steps=2, per device eval batch size=1, weight decay=0.01, optimizer=adamw\_8bit, and learning rate scheduler type=cosine.


After SFT, we quantized the fine-tuned model into Q5\_K\_M. We then evaluated the quantized model on the STEM gold and revision subsets of the HLE-verified dataset \cite{zhai_hle-verified_2026} as a held-out external test set (N=1,309). We selected this benchmark given its structural similarity to our dataset. HLE-verified was preferred over the original HLE benchmark because it provides verified, higher-accuracy questions and answers \cite{zhai_hle-verified_2026}. Response generation followed the same prompt used during post-training (Fig. S3), and grading used the same prompt and sampling parameters as in the model evaluation stage. We then compared performance between the baseline and post-trained versions of Qwen 3.5 9B Q5\_K\_M. All fine-tuning was performed on an RTX PRO 6000 GPU.


\subsection{Statistical analysis}

After conducting normality tests on the distributions of model response confidence scores for correct and incorrect answers (p>0.05), the Mann-Whiteney U test was performed at $\alpha$=0.05. The p-values for the models were corrected for multiple comparisons (Bonferroni) to determine statistical significance. The effect of post-training, i.e., changes in Correct and Incorrect response from post-training vs baseline, was assessed using McNemar’s test at $\alpha$=0.05.

\section{RESULTS}

\subsection{Annotator, reviewer, and team lead statistics}

Our recruitment process yielded a total of 241 annotators, reviewers, and team leads (Table S2). 80\% of them earned PhD in their expertise.

\subsection{Dataset generation results}

The data generation process yielded 398 examples, balanced over the physics, chemistry, mathematics, and biology domains (Fig. 2A). The taxonomy distributions (Fig. 2B) demonstrated that the L1 taxonomy categories were well distributed in Physics and Chemistry, but the examples from Mathematics and Biology were concentrated under discrete mathematics (49\%) and molecular biology (70\%) at L1 taxonomy, respectively. These unbalanced L1 taxonomy volumes in these domains were attributable to expertise represented in the annotator pool. Additionally, in Biology, the skewness of L1 taxonomy distribution toward molecular biology was attributable to the broad conceptual scope of molecular biology, which encompasses many mechanisms involving genes, RNA, proteins, enzymes, and cellular signaling. However, the L2 taxonomy in the largest L1 taxonomy categories was well distributed in all domains. Of note, many of the L2 topics in Biology appeared only once, and they were collapsed into the “other” category. Thus, the large “other” fraction in Biology should be interpreted as reflecting a diverse set of singleton L2 categories rather than a single homogeneous topic.

\begin{figure}[htbp]
\centering
\includegraphics[width=0.85\linewidth]{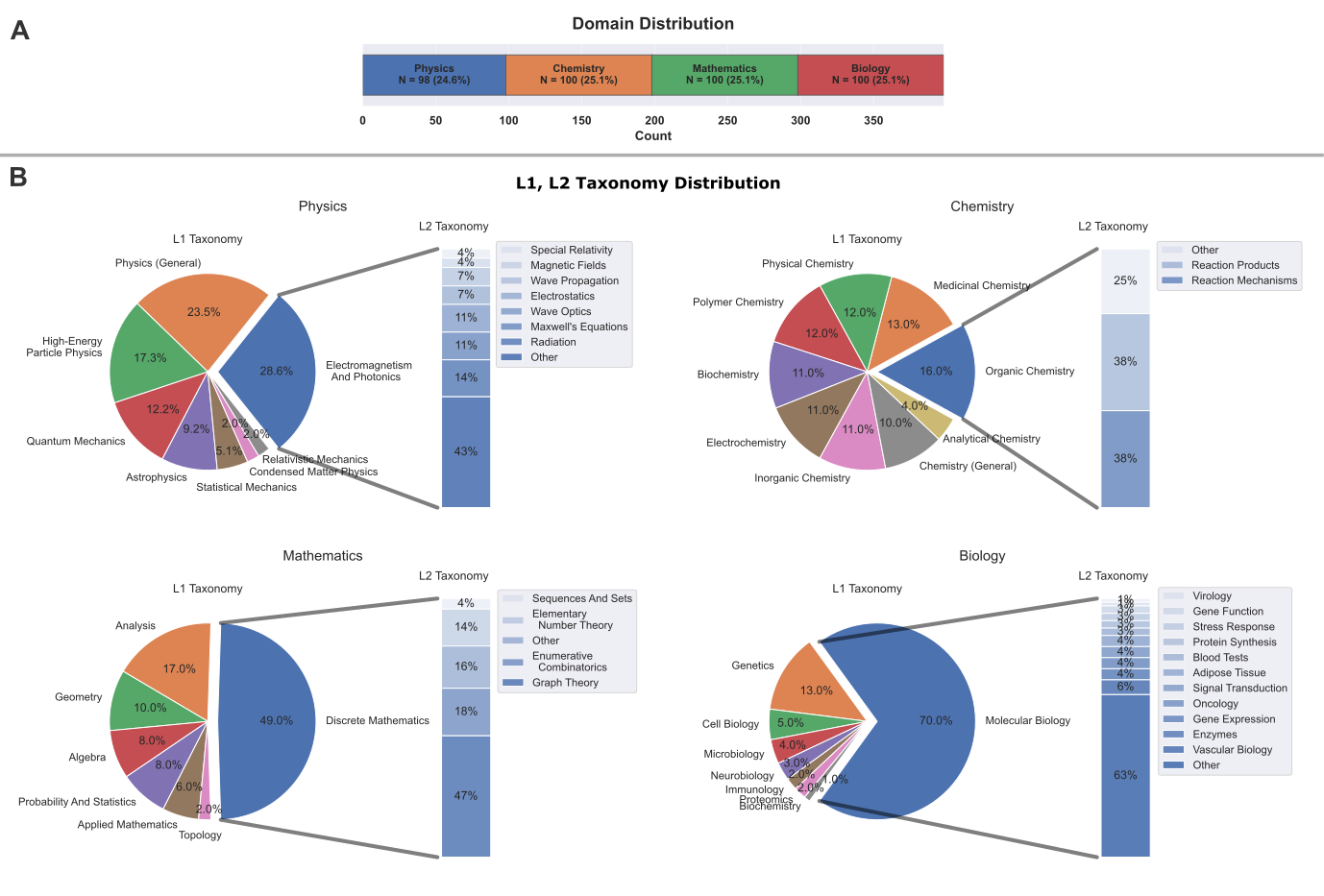}
\caption{Distributions of domains (A) and taxonomy (B). L2 taxonomy for the largest L1 taxonomy category shown in the colorbar. Examples with N=1 were designated as ‘other’ L2 taxonomy categories.}
\label{fig:taxonomy_distribution}
\end{figure}

\FloatBarrier

Token count distribution showed upper long tails in token counts for questions and response primarily driven by examples from Chemistry (Fig. 3A). Overall, the dataset exhibited domain-specific token distribution patterns, reflecting unique aspects across the domains. Token counts for questions in Mathematics were relatively lower than the other domains whereas token counts for combined response (explanation and ground truth answer) were comparable to examples from Physics and Chemistry (Fig. 3). The token counts for combined response in Biology were low relative to the other domains (Fig. 3B).

\begin{figure}[htbp]
\centering
\includegraphics[width=0.90\linewidth]{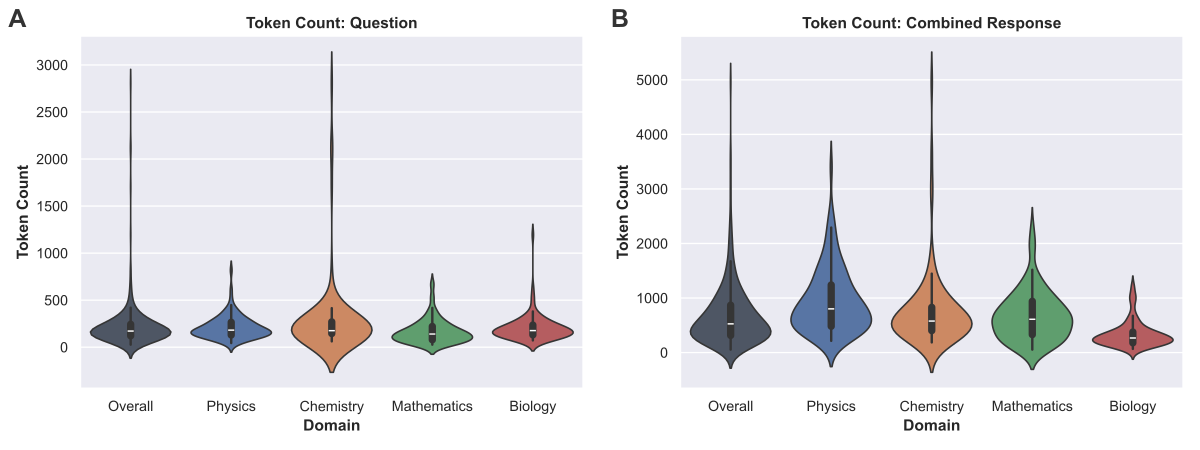}
\caption{Token distributions of questions (A) and combined response (explanation and answer; B) across the overall dataset and the four domains (from left to right).}
\label{fig:token_counts}
\end{figure}

\FloatBarrier

\subsection{Model performance on the dataset}

All 10 models scored below 25\%, indicating substantial headroom for improvement among current models (Fig. 4A). The proprietary models outperformed open-source models overall. Across all the models, higher reasoning efforts yielded higher scores, demonstrating both that the dataset demands multi-step reasoning and that it is sensitive to the degree of reasoning effort applied. Among open-source models, larger model size was associated with higher scores. At the domain level, the proprietary models scored relatively uniformly across domains whereas the open source models (Qwen 3.5) showed relatively weak performance in Chemistry, suggesting the dataset can inform decisions about which domains warrant additional training data.


Despite low scores, the models were generally overconfident, with median confidence exceeding 80\% with the exception of GPT 5.2 at medium reasoning effort, suggesting that models tend to hallucinate with high expressed confidence. (Fig. 4B). By comparison, GPT models expressed lower answer confidence than the other model families. Comparing confidence for correct and incorrect answers, the models showed more pronounced lower-tail distributions for incorrect answers than for correct answers except GPT 5.2 medium reasoning efforts. Statistical comparisons yielded mixed results: statistical significance for Claude Opus 4.5 high reasoning and Qwen 3.5 9B reasoning off at Bonferroni-corrected p$<$0.05; marginal significance for GPT 5.2 high reasoning effort and Qwen 3.5 27B reasoning enabled at uncorrected p$<$0.05 (Mann-Whitney U).



\begin{figure}[htbp]
\centering
\includegraphics[width=0.90\linewidth]{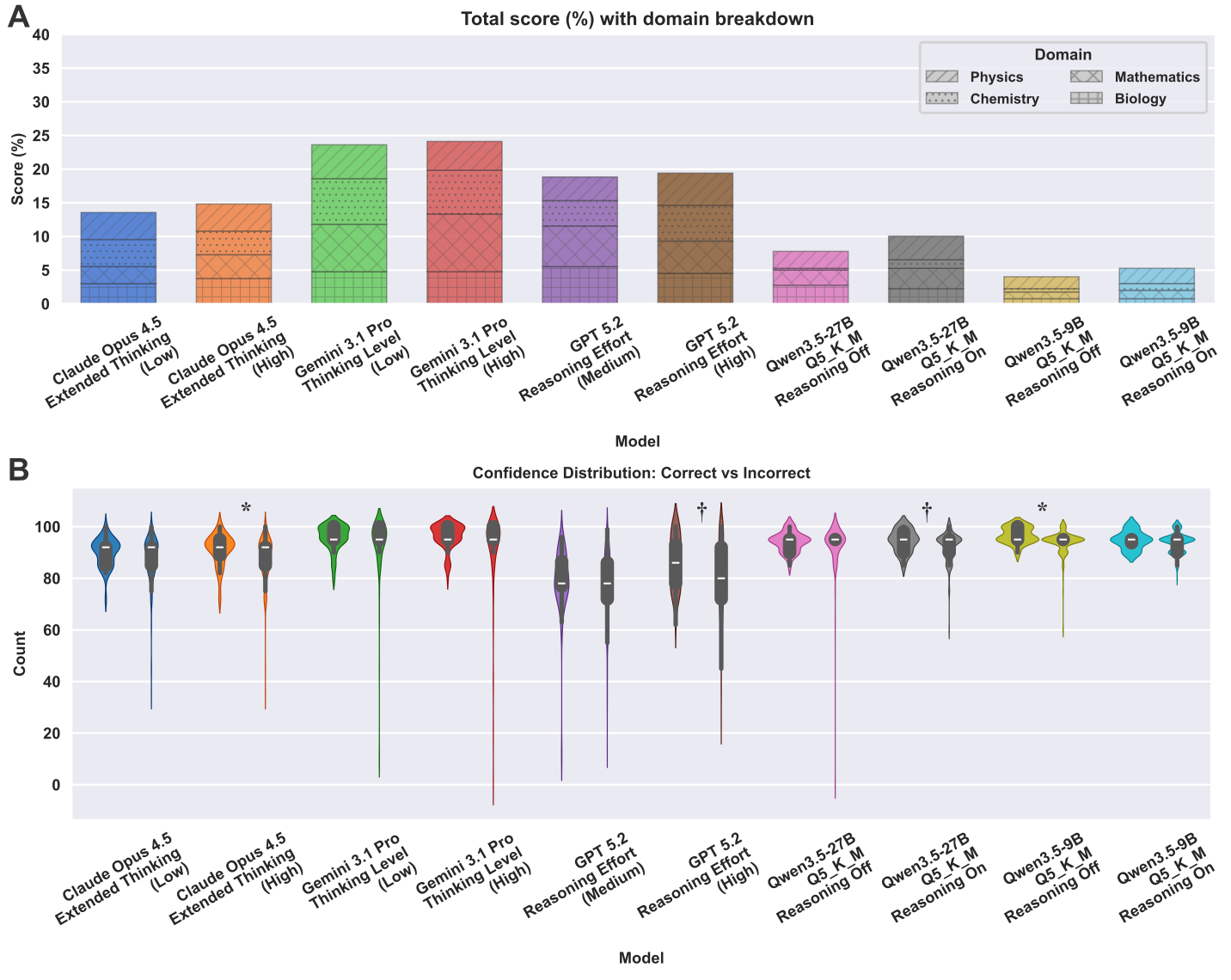}
\caption{Model scores and response confidence. A: Total score along with domain-specific scores (Physics, Chemistry, Mathematics, and Biology from top to bottom) of proprietary (left) and open source models on the dataset (right) with variations of reasoning efforts. B: Distribution of confidence of correct (left pair) and incorrect (right pair) model response. *, + represent Bonferroni-corrected p$<$0.05, uncorrected p$<$0.05 (Mann-Whitney U), respectively.}

\label{fig:model_comparison}
\end{figure}

\FloatBarrier

Upon finding that the completion token count distributions for the open-source models were bimodal (Fig. S5), we examined the reasoning traces of the open-source reasoning models, to which we had direct access. Analysis revealed that, when confronted with challenging questions, the models exhibited signs of uncertainty and entered extended reasoning loops that exhausted the maximum completion token budget (Fig. S5). We also found that the models appeared to retrieve information from training memory rather than engaging in generative multi-step reasoning (Fig. S5, \textit{Example 2}).


Pass rate distributions across 8 trials for the two Qwen 3.5 reasoning models (Fig. 5A) indicate that both models consistently gave incorrect answers, with the majority of pass rates falling below 4/8. Pass@k curves for the 27B model were consistently higher than those of the 9B model across all 8 trials (Fig. 5B). However, neither curve exhibited a steep elbow, suggesting that a prohibitively large number of attempts would be required to approach each model's performance ceiling.


\begin{figure}[htbp]
\centering
\includegraphics[width=0.80\linewidth]{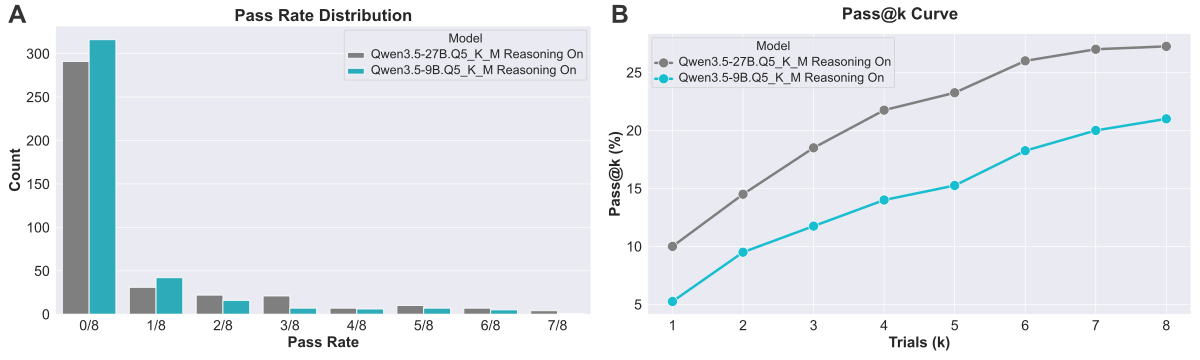}
\caption{Pass rate distribution (A) and pass@k curves (B) over over 8 trials for the two open source reasoning models: Qwen 3.5 27B Q5\_K\_M and Qwen 3.5 9B Q5\_K\_M.}
\label{fig:pass_rate}
\end{figure}

\FloatBarrier

\subsection{Post-training results}

Fine-tuning Qwen 3.5 9B Q5\_K\_M on the separate proprietary dataset (N=2,000) improved overall scores on the HLE-verified STEM subset by 15\% of the baseline performance (Fig. 6A). The improvement was statistically significant (p=0.045; McNemar’s Test). This gain was driven primarily by improvement on the verifiable question-and-answer subset (22\% of the baseline; p=0.034) rather than the multiple-choice-question subset (5.8\% of the baseline; Fig. 6). However, the effect size was moderate. Further, though the effect of post-training on overall scores was statistically significant, a notable number of examples that were answered correctly at baseline were answered incorrectly after fine-tuning (Tables S3-S5), suggesting that more targeted post-training strategies could yield further gains (refer to the limitations and future directions section for more discussion).


\begin{figure}[htbp]
\centering
\includegraphics[width=0.90\linewidth]{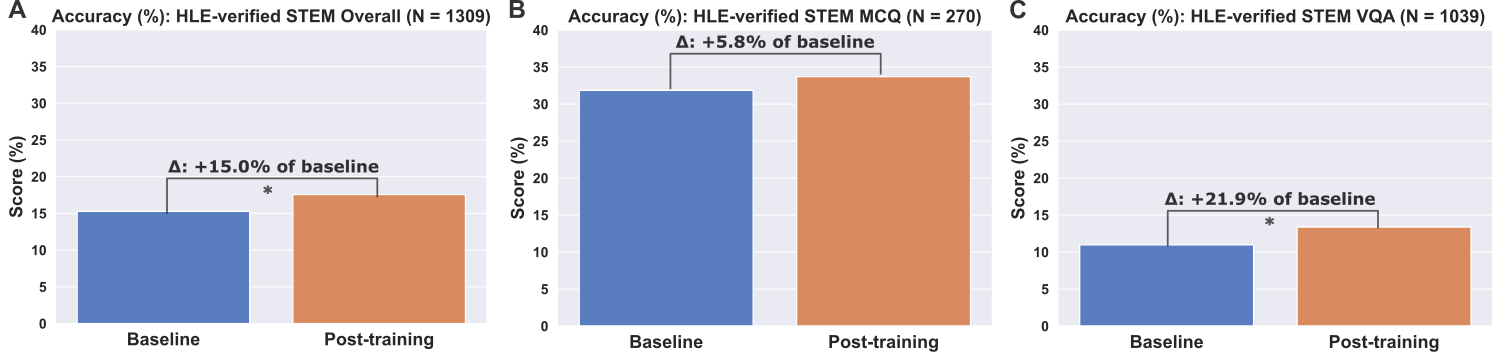}
\caption{Scores of post-trained vs baseline Qwen 3.5 9B Reasoning Off for HLE-verified STEM overall (A), multiple choice question (MCQ) subset (B), and verifiable question and answer (VQA) subset (C). * represent p$<$0.05 (McNemar’s Test). The baseline and post-trained Qwen 3.5 9B models were quantized at Q5\_K\_M for the scoring.}
\label{fig:posttraining_effect}
\end{figure}

\FloatBarrier

\section{DISCUSSIONS}

In summary, we presented ‘Expert-validated STEM QA’ (N=398) with careful dataset design and rigorous consensus-driven review process from 241 domain experts (80\% of them earned PhD). The frontier models and open source models scored lower than 25\% of the total scores. Post-training of the open source model on the separate, private version of the dataset (N=2,000) increased its score by 15\% of the baseline model performance (p=0.045).

\subsection{Highlights of the dataset}

Accuracy of datasets in the scientific domain is critical and often difficult to define, as the correct answer at present may not hold in the future. Moreover, correctness depends on context and the audience. For example, the answer "blue" to "What color is the sky?" carries different validity depending on who is asked. A layperson would accept "blue" as correct. A detail-oriented person might respond: "The context is incomplete. \textit{In the morning}, the sky is light blue, but orange in the evening." A STEM expert would expect a more sophisticated answer: "The sky appears light blue due to \textit{light scattering} during daytime, but it has \textit{no intrinsic color}." A STEM benchmark without proper expert validation therefore weakens its reliability for model evaluation. Our multi-layer, consensus-based review process is aligned with current efforts emphasizing data quality in the field \cite{zhai_hle-verified_2026, rein_gpqa_2023}.


Our dataset was generated with verified search-proof evidence, and a larger portion is held private to mitigate data leakage — a critical factor for objectively measuring model generalizability. Our analysis of model reasoning processes indicates that models sometimes draw on internally memorized material (Fig. S6), making leakage prevention especially important for reasoning models. Datasets curated from public sources are more susceptible to leakage; consequently, search-proof benchmarks like GPQA Diamond \cite{rein_gpqa_2023} remain valuable even as model performance on some of them nears saturation. A larger, private version of our dataset is available for leakage-proof evaluation upon request.


Additional strengths include substantial headroom for model improvement and expert verification of responses, which guards against questions designed to exploit model weaknesses — as seen in earlier versions of HLE \cite{phan_humanitys_2025} — in favor of questions requiring genuine multi-step reasoning.


We also demonstrated the potential utility of our dataset for model training through post-training experiments (Fig. 6; Tables S3–S5). Note that this serves as an illustrative example; more sophisticated reasoning strategies and larger data volumes would be required in practice (refer to the Limitations and future directions section). Several existing datasets \cite{yuan_naturalreasoning_2025, moshkov_aimo-2_2025, noauthor_nvidiaopensciencereasoning-2_2026} are intended for post-training and are generally sourced from curated public or synthetic data. We have a separate, larger, human-expert-validated dataset available to support model improvement via pre-, mid-, and post-training.


\subsection{Findings in relation to prior literature}

Evaluation of model performance on our datasets (Figs. 4–5, S5, S6) reveals that assessment should be holistic, extending beyond answer accuracy alone \cite{liang_holistic_2023}. Based on our findings, we highlight two key observations: model hallucination and reasoning efficiency.


Models consistently assigned high confidence scores to their answers even when incorrect (Fig. 4), a well-documented phenomenon of hallucination in LLMs \cite{alansari_large_2026, li_mitigating_2025}. This pattern is particularly consequential in scientific contexts, where answer accuracy is paramount. STEM benchmarks should therefore maintain rigorous standards of answer validity to serve as meaningful probes of model hallucination.

Further analysis of reasoning traces revealed that models sometimes entered repetitive reasoning loops without meaningful progression (Fig. S6). While this remains an empirical observation, it motivates systematic investigation into reasoning efficiency and depth in STEM settings — for instance, through metrics such as deep-thinking ratio \cite{chen_think_2026}.


\subsection{Limitations and future directions}

One of the limitations of this study is that we were unable to systematically evaluate the model's reasoning processes on our dataset as access to reasoning processes varies with models, especially proprietary models. Secondly, our data volume (N=398) is lower than the synthetic and curated dataset (often millions) due to efforts from human experts to create such data. We have more data available in the private version, and careful model training strategy would aid to compensate for low data volume. Another limitation of the study is we performed SFT on the non-reasoning model. SFT on a reasoning model was challenging as reasoning traces were unavailable at this time, and SFT is known to be susceptible to catastrophic forgetting. Thus, the interpretation of our results is rather limited.

Our future direction includes systematic assessment of the reasoning process for better model evaluation, conducting more sophisticated post-training, e.g., reinforcement learning with data mixture, and evaluating more recent models on our dataset.

In conclusion, we presented here expert-validated STEM QA. We believe that our work would help advance the research community for AI in STEM.

\newpage

\begin{ack}
The authors would like to thank Amit Kumar Jha for his assistance with project document collection and Danny Arlen de Jesus Gómez Ramírez for assistance with creating the taxonomy. The authors would also like to thank contributors for creating the dataset. 
\end{ack}


\bibliographystyle{plain} 
\bibliography{references} 

@misc{liang_holistic_2023,
	title = {Holistic {Evaluation} of {Language} {Models}},
	url = {http://arxiv.org/abs/2211.09110},
	doi = {10.48550/arXiv.2211.09110},
	urldate = {2026-05-06},
	publisher = {arXiv},
	author = {Liang, Percy and Bommasani, Rishi and Lee, Tony and Tsipras, Dimitris and Soylu, Dilara and Yasunaga, Michihiro and Zhang, Yian and Narayanan, Deepak and Wu, Yuhuai and Kumar, Ananya and Newman, Benjamin and Yuan, Binhang and Yan, Bobby and Zhang, Ce and Cosgrove, Christian and Manning, Christopher D. and Ré, Christopher and Acosta-Navas, Diana and Hudson, Drew A. and Zelikman, Eric and Durmus, Esin and Ladhak, Faisal and Rong, Frieda and Ren, Hongyu and Yao, Huaxiu and Wang, Jue and Santhanam, Keshav and Orr, Laurel and Zheng, Lucia and Yuksekgonul, Mert and Suzgun, Mirac and Kim, Nathan and Guha, Neel and Chatterji, Niladri and Khattab, Omar and Henderson, Peter and Huang, Qian and Chi, Ryan and Xie, Sang Michael and Santurkar, Shibani and Ganguli, Surya and Hashimoto, Tatsunori and Icard, Thomas and Zhang, Tianyi and Chaudhary, Vishrav and Wang, William and Li, Xuechen and Mai, Yifan and Zhang, Yuhui and Koreeda, Yuta},
	month = oct,
	year = {2023},
	note = {arXiv:2211.09110},
}

@misc{li_mitigating_2025,
	title = {Mitigating {Hallucination} in {Large} {Language} {Models} ({LLMs}): {An} {Application}-{Oriented} {Survey} on {RAG}, {Reasoning}, and {Agentic} {Systems}},
	shorttitle = {Mitigating {Hallucination} in {Large} {Language} {Models} ({LLMs})},
	url = {http://arxiv.org/abs/2510.24476},
	doi = {10.48550/arXiv.2510.24476},
	urldate = {2026-05-06},
	publisher = {arXiv},
	author = {Li, Yihan and Fu, Xiyuan and Verma, Ghanshyam and Buitelaar, Paul and Liu, Mingming},
	month = oct,
	year = {2025},
	note = {arXiv:2510.24476},
}

@misc{alansari_large_2026,
	title = {Large {Language} {Models} {Hallucination}: {A} {Comprehensive} {Survey}},
	shorttitle = {Large {Language} {Models} {Hallucination}},
	url = {http://arxiv.org/abs/2510.06265},
	doi = {10.48550/arXiv.2510.06265},
	urldate = {2026-05-06},
	publisher = {arXiv},
	author = {Alansari, Aisha and Luqman, Hamzah},
	month = mar,
	year = {2026},
	note = {arXiv:2510.06265},
}

@misc{yuan_naturalreasoning_2025,
	title = {{NaturalReasoning}: {Reasoning} in the {Wild} with 2.{8M} {Challenging} {Questions}},
	shorttitle = {{NaturalReasoning}},
	url = {http://arxiv.org/abs/2502.13124},
	doi = {10.48550/arXiv.2502.13124},
	urldate = {2026-05-06},
	publisher = {arXiv},
	author = {Yuan, Weizhe and Yu, Jane and Jiang, Song and Padthe, Karthik and Li, Yang and Kulikov, Ilia and Cho, Kyunghyun and Wang, Dong and Tian, Yuandong and Weston, Jason E. and Li, Xian},
	month = nov,
	year = {2025},
	note = {arXiv:2502.13124},
}

@misc{noauthor_futurehouseether0-benchmark_2025,
	title = {futurehouse/ether0-benchmark · {Datasets} at {Hugging} {Face}},
	url = {https://huggingface.co/datasets/futurehouse/ether0-benchmark},
	urldate = {2026-05-06},
	month = jun,
	year = {2025},
}

@misc{noauthor_nvidiaopensciencereasoning-2_2026,
	title = {nvidia/{OpenScienceReasoning}-2 · {Datasets} at {Hugging} {Face}},
	url = {https://huggingface.co/datasets/nvidia/OpenScienceReasoning-2},
	urldate = {2026-05-06},
	month = feb,
	year = {2026},
}

@misc{glazer_frontiermath_2025,
	title = {{FrontierMath}: {A} {Benchmark} for {Evaluating} {Advanced} {Mathematical} {Reasoning} in {AI}},
	shorttitle = {{FrontierMath}},
	url = {http://arxiv.org/abs/2411.04872},
	doi = {10.48550/arXiv.2411.04872},
	urldate = {2026-05-06},
	publisher = {arXiv},
	author = {Glazer, Elliot and Erdil, Ege and Besiroglu, Tamay and Chicharro, Diego and Chen, Evan and Gunning, Alex and Olsson, Caroline Falkman and Denain, Jean-Stanislas and Ho, Anson and Santos, Emily de Oliveira and Järviniemi, Olli and Barnett, Matthew and Sandler, Robert and Vrzala, Matej and Sevilla, Jaime and Ren, Qiuyu and Pratt, Elizabeth and Levine, Lionel and Barkley, Grant and Stewart, Natalie and Grechuk, Bogdan and Grechuk, Tetiana and Enugandla, Shreepranav Varma and Wildon, Mark},
	month = dec,
	year = {2025},
	note = {arXiv:2411.04872},
}

@misc{moshkov_aimo-2_2025,
	title = {{AIMO}-2 {Winning} {Solution}: {Building} {State}-of-the-{Art} {Mathematical} {Reasoning} {Models} with {OpenMathReasoning} dataset},
	shorttitle = {{AIMO}-2 {Winning} {Solution}},
	url = {http://arxiv.org/abs/2504.16891},
	doi = {10.48550/arXiv.2504.16891},
	urldate = {2026-05-06},
	publisher = {arXiv},
	author = {Moshkov, Ivan and Hanley, Darragh and Sorokin, Ivan and Toshniwal, Shubham and Henkel, Christof and Schifferer, Benedikt and Du, Wei and Gitman, Igor},
	month = apr,
	year = {2025},
	note = {arXiv:2504.16891},
}

@misc{ma_sci-reason_2025,
	title = {{SCI}-{Reason}: {A} {Dataset} with {Chain}-of-{Thought} {Rationales} for {Complex} {Multimodal} {Reasoning} in {Academic} {Areas}},
	shorttitle = {{SCI}-{Reason}},
	url = {http://arxiv.org/abs/2504.06637},
	doi = {10.48550/arXiv.2504.06637},
	urldate = {2026-05-06},
	publisher = {arXiv},
	author = {Ma, Chenghao and E, Haihong and Ding, Junpeng and Zhang, Jun and Ma, Ziyan and Qing, Huang and Gao, Bofei and Chen, Liang and Zhu, Yifan and Song, Meina},
	month = sep,
	year = {2025},
	note = {arXiv:2504.06637},
}

@misc{lu_scp-116k_2025,
	title = {{SCP}-{116K}: {A} {High}-{Quality} {Problem}-{Solution} {Dataset} and a {Generalized} {Pipeline} for {Automated} {Extraction} in the {Higher} {Education} {Science} {Domain}},
	shorttitle = {{SCP}-{116K}},
	url = {http://arxiv.org/abs/2501.15587},
	doi = {10.48550/arXiv.2501.15587},
	urldate = {2026-05-06},
	publisher = {arXiv},
	author = {Lu, Dakuan and Tan, Xiaoyu and Xu, Rui and Yao, Tianchu and Qu, Chao and Chu, Wei and Xu, Yinghui and Qi, Yuan},
	month = aug,
	year = {2025},
	note = {arXiv:2501.15587},
}

@misc{li_system_2025,
	title = {From {System} 1 to {System} 2: {A} {Survey} of {Reasoning} {Large} {Language} {Models}},
	shorttitle = {From {System} 1 to {System} 2},
	url = {http://arxiv.org/abs/2502.17419},
	doi = {10.48550/arXiv.2502.17419},
	urldate = {2026-05-06},
	publisher = {arXiv},
	author = {Li, Zhong-Zhi and Zhang, Duzhen and Zhang, Ming-Liang and Zhang, Jiaxin and Liu, Zengyan and Yao, Yuxuan and Xu, Haotian and Zheng, Junhao and Wang, Pei-Jie and Chen, Xiuyi and Zhang, Yingying and Yin, Fei and Dong, Jiahua and Li, Zhiwei and Bi, Bao-Long and Mei, Ling-Rui and Fang, Junfeng and Liang, Xiao and Guo, Zhijiang and Song, Le and Liu, Cheng-Lin},
	month = jun,
	year = {2025},
	note = {arXiv:2502.17419},
}

@article{abramson_accurate_2024,
	title = {Accurate structure prediction of biomolecular interactions with {AlphaFold} 3},
	volume = {630},
	copyright = {2024 The Author(s)},
	issn = {1476-4687},
	url = {https://www.nature.com/articles/s41586-024-07487-w},
	doi = {10.1038/s41586-024-07487-w},
	language = {en},
	number = {8016},
	urldate = {2026-05-06},
	journal = {Nature},
	author = {Abramson, Josh and Adler, Jonas and Dunger, Jack and Evans, Richard and Green, Tim and Pritzel, Alexander and Ronneberger, Olaf and Willmore, Lindsay and Ballard, Andrew J. and Bambrick, Joshua and Bodenstein, Sebastian W. and Evans, David A. and Hung, Chia-Chun and O’Neill, Michael and Reiman, David and Tunyasuvunakool, Kathryn and Wu, Zachary and Žemgulytė, Akvilė and Arvaniti, Eirini and Beattie, Charles and Bertolli, Ottavia and Bridgland, Alex and Cherepanov, Alexey and Congreve, Miles and Cowen-Rivers, Alexander I. and Cowie, Andrew and Figurnov, Michael and Fuchs, Fabian B. and Gladman, Hannah and Jain, Rishub and Khan, Yousuf A. and Low, Caroline M. R. and Perlin, Kuba and Potapenko, Anna and Savy, Pascal and Singh, Sukhdeep and Stecula, Adrian and Thillaisundaram, Ashok and Tong, Catherine and Yakneen, Sergei and Zhong, Ellen D. and Zielinski, Michal and Žídek, Augustin and Bapst, Victor and Kohli, Pushmeet and Jaderberg, Max and Hassabis, Demis and Jumper, John M.},
	month = jun,
	year = {2024},
	pages = {493--500},
}

@misc{noauthor_qwen35_2026,
	title = {Qwen3.5 - {How} to {Run} {Locally} {\textbar} {Unsloth} {Documentation}},
	url = {https://unsloth.ai/docs/models/qwen3.5},
	language = {en},
	urldate = {2026-05-06},
	month = apr,
	year = {2026},
}

@misc{noauthor_unslothqwen35-9b-gguf_2026,
	title = {unsloth/{Qwen3}.5-{9B}-{GGUF} · {Hugging} {Face}},
	url = {https://huggingface.co/unsloth/Qwen3.5-9B-GGUF},
	urldate = {2026-05-06},
	month = apr,
	year = {2026},
}

@misc{noauthor_unslothqwen35-27b-gguf_2026,
	title = {unsloth/{Qwen3}.5-{27B}-{GGUF} · {Hugging} {Face}},
	url = {https://huggingface.co/unsloth/Qwen3.5-27B-GGUF},
	urldate = {2026-05-06},
	month = apr,
	year = {2026},
}

@misc{anthropic_claude_2025,
	type = {Large language model},
	title = {Claude {Opus} 4.5},
	url = {claude.ai},
	author = {Anthropic},
	month = nov,
	year = {2025},
}

@misc{openai_chatgpt_2025,
	type = {[{Large} language model]},
	title = {{ChatGPT} ({GPT}-5.2), {Dec} 12, 2025},
	url = {https://chat.openai.com},
	author = {OpenAI},
	month = dec,
	year = {2025},
}

@misc{google_gemini_2026,
	type = {Large language model.},
	title = {Gemini 3.1 {Pro} {Preview}},
	url = {https://gemini.google.com/},
	author = {Google},
	month = feb,
	year = {2026},
}

@misc{qwen_team_qwen35_2026,
	title = {Qwen3.5: {Towards} {Native} {Multimodal} {Agents}},
	url = {https://qwen.ai/blog?id=qwen3.5},
	author = {{Qwen Team}},
	month = feb,
	year = {2026},
}

@misc{daniel_han_unsloth_2023,
	title = {Unsloth},
	url = {https://github.com/unslothai/unsloth},
	author = {Daniel Han, Michael Han and team, Unsloth},
	year = {2023},
}

@misc{zhai_hle-verified_2026,
	title = {{HLE}-{Verified}: {A} {Systematic} {Verification} and {Structured} {Revision} of {Humanity}'s {Last} {Exam}},
	shorttitle = {{HLE}-{Verified}},
	url = {http://arxiv.org/abs/2602.13964},
	doi = {10.48550/arXiv.2602.13964},
	urldate = {2026-05-05},
	publisher = {arXiv},
	author = {Zhai, Weiqi and Wang, Zhihai and Wang, Jinghang and Yang, Boyu and Li, Xiaogang and Xu, Xander and Wang, Bohan and Wang, Peng and Wu, Xingzhe and Li, Anfeng and Feng, Qiyuan and Zhou, Yuhao and Han, Shoulin and Luo, Wenjie and Li, Yiyuan and Wang, Yaxuan and Luo, Ruixian and Lin, Guojie and Xiao, Peiyao and Xu, Chengliang and Wang, Ben and Wang, Zeyu and Chen, Zichao and Ye, Jianan and Hu, Yijie and Chen, Jialong and Shen, Zongwen and Xu, Yuliang and Yang, An and Yu, Bowen and Liu, Dayiheng and Lin, Junyang and Wei, Hu and Shen, Que and Zhao, Bing},
	month = feb,
	year = {2026},
	note = {arXiv:2602.13964},
}

@misc{chen_think_2026,
	title = {Think {Deep}, {Not} {Just} {Long}: {Measuring} {LLM} {Reasoning} {Effort} via {Deep}-{Thinking} {Tokens}},
	shorttitle = {Think {Deep}, {Not} {Just} {Long}},
	url = {http://arxiv.org/abs/2602.13517},
	doi = {10.48550/arXiv.2602.13517},
	urldate = {2026-05-05},
	publisher = {arXiv},
	author = {Chen, Wei-Lin and Peng, Liqian and Tan, Tian and Zhao, Chao and Chen, Blake JianHang and Lin, Ziqian and Go, Alec and Meng, Yu},
	month = feb,
	year = {2026},
	note = {arXiv:2602.13517},
}

@article{botelho_scale_2025,
	title = {Scale dichotomization reduces customer racial discrimination and income inequality},
	volume = {639},
	copyright = {2025 The Author(s)},
	issn = {1476-4687},
	url = {https://www.nature.com/articles/s41586-025-08599-7},
	doi = {10.1038/s41586-025-08599-7},
	language = {en},
	number = {8054},
	urldate = {2025-10-30},
	journal = {Nature},
	author = {Botelho, Tristan L. and Jun, Sora and Humes, Demetrius and DeCelles, Katherine A.},
	month = mar,
	year = {2025},
	pages = {395--403},
}

@misc{team_supergpqa_2025,
	title = {{SuperGPQA}: {Scaling} {LLM} {Evaluation} across 285 {Graduate} {Disciplines}},
	shorttitle = {{SuperGPQA}},
	url = {http://arxiv.org/abs/2502.14739},
	doi = {10.48550/arXiv.2502.14739},
	urldate = {2025-10-30},
	publisher = {arXiv},
	author = {Team, M.-A.-P. and Du, Xinrun and Yao, Yifan and Ma, Kaijing and Wang, Bingli and Zheng, Tianyu and Zhu, King and Liu, Minghao and Liang, Yiming and Jin, Xiaolong and Wei, Zhenlin and Zheng, Chujie and Deng, Kaixin and Gavin, Shawn and Jia, Shian and Jiang, Sichao and Liao, Yiyan and Li, Rui and Li, Qinrui and Li, Sirun and Li, Yizhi and Li, Yunwen and Ma, David and Ni, Yuansheng and Que, Haoran and Wang, Qiyao and Wen, Zhoufutu and Wu, Siwei and Hsing, Tyshawn and Xu, Ming and Yang, Zhenzhu and Wang, Zekun Moore and Zhou, Junting and Bai, Yuelin and Bu, Xingyuan and Cai, Chenglin and Chen, Liang and Chen, Yifan and Cheng, Chengtuo and Cheng, Tianhao and Ding, Keyi and Huang, Siming and Huang, Yun and Li, Yaoru and Li, Yizhe and Li, Zhaoqun and Liang, Tianhao and Lin, Chengdong and Lin, Hongquan and Ma, Yinghao and Pang, Tianyang and Peng, Zhongyuan and Peng, Zifan and Qi, Qige and Qiu, Shi and Qu, Xingwei and Quan, Shanghaoran and Tan, Yizhou and Wang, Zili and Wang, Chenqing and Wang, Hao and Wang, Yiya and Wang, Yubo and Xu, Jiajun and Yang, Kexin and Yuan, Ruibin and Yue, Yuanhao and Zhan, Tianyang and Zhang, Chun and Zhang, Jinyang and Zhang, Xiyue and Zhang, Xingjian and Zhang, Yue and Zhao, Yongchi and Zheng, Xiangyu and Zhong, Chenghua and Gao, Yang and Li, Zhoujun and Liu, Dayiheng and Liu, Qian and Liu, Tianyu and Ni, Shiwen and Peng, Junran and Qin, Yujia and Su, Wenbo and Wang, Guoyin and Wang, Shi and Yang, Jian and Yang, Min and Cao, Meng and Yue, Xiang and Zhang, Zhaoxiang and Zhou, Wangchunshu and Liu, Jiaheng and Lin, Qunshu and Huang, Wenhao and Zhang, Ge},
	month = mar,
	year = {2025},
	note = {arXiv:2502.14739},
}

@misc{mallinar_scalable_2025,
	title = {A {Scalable} {Framework} for {Evaluating} {Health} {Language} {Models}},
	url = {http://arxiv.org/abs/2503.23339},
	doi = {10.48550/arXiv.2503.23339},
	urldate = {2025-08-28},
	publisher = {arXiv},
	author = {Mallinar, Neil and Heydari, A. Ali and Liu, Xin and Faranesh, Anthony Z. and Winslow, Brent and Hammerquist, Nova and Graef, Benjamin and Speed, Cathy and Malhotra, Mark and Patel, Shwetak and Prieto, Javier L. and McDuff, Daniel and Metwally, Ahmed A.},
	month = apr,
	year = {2025},
	note = {arXiv:2503.23339},
}

@misc{rein_gpqa_2023,
	title = {{GPQA}: {A} {Graduate}-{Level} {Google}-{Proof} {Q}\&{A} {Benchmark}},
	shorttitle = {{GPQA}},
	url = {http://arxiv.org/abs/2311.12022},
	doi = {10.48550/arXiv.2311.12022},
	urldate = {2025-01-29},
	publisher = {arXiv},
	author = {Rein, David and Hou, Betty Li and Stickland, Asa Cooper and Petty, Jackson and Pang, Richard Yuanzhe and Dirani, Julien and Michael, Julian and Bowman, Samuel R.},
	month = nov,
	year = {2023},
	note = {arXiv:2311.12022},
}

@misc{phan_humanitys_2025,
	title = {Humanity's {Last} {Exam}},
	copyright = {Creative Commons Attribution 4.0 International},
	url = {https://arxiv.org/abs/2501.14249},
	doi = {10.48550/ARXIV.2501.14249},
	urldate = {2025-01-27},
	publisher = {arXiv},
	author = {Phan, Long and Gatti, Alice and Han, Ziwen and Li, Nathaniel and Hu, Josephina and Zhang, Hugh and Shi, Sean and Choi, Michael and Agrawal, Anish and Chopra, Arnav and Khoja, Adam and Kim, Ryan and Hausenloy, Jason and Zhang, Oliver and Mazeika, Mantas and Anderson, Daron and Nguyen, Tung and Mahmood, Mobeen and Feng, Fiona and Feng, Steven Y. and Zhao, Haoran and Yu, Michael and Gangal, Varun and Zou, Chelsea and Wang, Zihan and Wang, Jessica P. and Kumar, Pawan and Pokutnyi, Oleksandr and Gerbicz, Robert and Popov, Serguei and Levin, John-Clark and Kazakov, Mstyslav and Schmitt, Johannes and Galgon, Geoff and Sanchez, Alvaro and Lee, Yongki and Yeadon, Will and Sauers, Scott and Roth, Marc and Agu, Chidozie and Riis, Søren and Giska, Fabian and Utpala, Saiteja and Giboney, Zachary and Goshu, Gashaw M. and Xavier, Joan of Arc and Crowson, Sarah-Jane and Naiya, Mohinder Maheshbhai and Burns, Noah and Finke, Lennart and Cheng, Zerui and Park, Hyunwoo and Fournier-Facio, Francesco and Wydallis, John and Nandor, Mark and Singh, Ankit and Gehrunger, Tim and Cai, Jiaqi and McCarty, Ben and Duclosel, Darling and Nam, Jungbae and Zampese, Jennifer and Hoerr, Ryan G. and Bacho, Aras and Loume, Gautier Abou and Galal, Abdallah and Cao, Hangrui and Garretson, Alexis C and Sileo, Damien and Ren, Qiuyu and Cojoc, Doru and Arkhipov, Pavel and Qazi, Usman and Li, Lianghui and Motwani, Sumeet and de Witt, Christian Schroeder and Taylor, Edwin and Veith, Johannes and Singer, Eric and Hartman, Taylor D. and Rissone, Paolo and Jin, Jaehyeok and Shi, Jack Wei Lun and Willcocks, Chris G. and Robinson, Joshua and Mikov, Aleksandar and Prabhu, Ameya and Tang, Longke and Alapont, Xavier and Uro, Justine Leon and Zhou, Kevin and Santos, Emily de Oliveira and Maksimov, Andrey Pupasov and Vendrow, Edward and Zenitani, Kengo and Guillod, Julien and Li, Yuqi and Vendrow, Joshua and Kuchkin, Vladyslav and Ze-An, Ng and Marion, Pierre and Efremov, Denis and Lynch, Jayson and Liang, Kaiqu and Gritsevskiy, Andrew and Martinez, Dakotah and Pageler, Ben and Crispino, Nick and Zvonkine, Dimitri and Fraga, Natanael Wildner and Soori, Saeed and Press, Ori and Tang, Henry and Salazar, Julian and Green, Sean R. and Brüssel, Lina and Twayana, Moon and Dieuleveut, Aymeric and Rogers, T. Ryan and Zhang, Wenjin and Li, Bikun and Yang, Jinzhou and Rao, Arun and Loiseau, Gabriel and Kalinin, Mikhail and Lukas, Marco and Manolescu, Ciprian and Mishra, Subrata and Kamdoum, Ariel Ghislain Kemogne and Kreiman, Tobias and Hogg, Tad and Jin, Alvin and Bosio, Carlo and Sun, Gongbo and Coppola, Brian P and Tarver, Tim and Heidinger, Haline and Sayous, Rafael and Ivanov, Stefan and Cavanagh, Joseph M and Shen, Jiawei and Imperial, Joseph Marvin and Schwaller, Philippe and Senthilkuma, Shaipranesh and Bran, Andres M and Dehghan, Ali and Algaba, Andres and Verbeken, Brecht and Noever, David and P, Ragavendran and Schut, Lisa and Sucholutsky, Ilia and Zheltonozhskii, Evgenii and Lim, Derek and Stanley, Richard and Sivarajan, Shankar and Yang, Tong and Maar, John and Wykowski, Julian and Oller, Martí and Sandlin, Jennifer and Sahu, Anmol and Hu, Yuzheng and Fish, Sara and Heydari, Nasser and Apronti, Archimedes and Rawal, Kaivalya and Vilchis, Tobias Garcia and Zu, Yuexuan and Lackner, Martin and Koppel, James and Nguyen, Jeremy and Antonenko, Daniil S. and Chern, Steffi and Zhao, Bingchen and Arsene, Pierrot and Goldfarb, Alan and Ivanov, Sergey and Poświata, Rafał and Wang, Chenguang and Li, Daofeng and Crisostomi, Donato and Achilleos, Andrea and Myklebust, Benjamin and Sen, Archan and Perrella, David and Kaparov, Nurdin and Inlow, Mark H and Zang, Allen and Thornley, Elliott and Orel, Daniil and Poritski, Vladislav and Ben-David, Shalev and Berger, Zachary and Whitfill, Parker and Foster, Michael and Munro, Daniel and Ho, Linh and Hava, Dan Bar and Kuchkin, Aleksey and Lauff, Robert and Holmes, David and Sommerhage, Frank and Schneider, Keith and Kazibwe, Zakayo and Stambaugh, Nate and Singh, Mukhwinder and Magoulas, Ilias and Clarke, Don and Kim, Dae Hyun and Dias, Felipe Meneguitti and Elser, Veit and Agarwal, Kanu Priya and Vilchis, Victor Efren Guadarrama and Klose, Immo and Demian, Christoph and Anantheswaran, Ujjwala and Zweiger, Adam and Albani, Guglielmo and Li, Jeffery and Daans, Nicolas and Radionov, Maksim and Rozhoň, Václav and Ma, Ziqiao and Stump, Christian and Berkani, Mohammed and Platnick, Jacob and Nevirkovets, Volodymyr and Basler, Luke and Piccardo, Marco and Jeanplong, Ferenc and Cohen, Niv and Tkadlec, Josef and Rosu, Paul and Padlewski, Piotr and Barzowski, Stanislaw and Montgomery, Kyle and Menezes, Aline and Patel, Arkil and Wang, Zixuan and Tucker-Foltz, Jamie and Stade, Jack and Goertzen, Tom and Kazemi, Fereshteh and Milbauer, Jeremiah and Ambay, John Arnold and Shukla, Abhishek and Labrador, Yan Carlos Leyva and Givré, Alan and Wolff, Hew and Rossbach, Vivien and Aziz, Muhammad Fayez and Kaddar, Younesse and Chen, Yanxu and Zhang, Robin and Pan, Jiayi and Terpin, Antonio and Muennighoff, Niklas and Schoelkopf, Hailey and Zheng, Eric and Carmi, Avishy and Jones, Adam and Shah, Jainam and Brown, Ethan D. L. and Zhu, Kelin and Bartolo, Max and Wheeler, Richard and Ho, Andrew and Barkan, Shaul and Wang, Jiaqi and Stehberger, Martin and Kretov, Egor and Sridhar, Kaustubh and EL-Wasif, Zienab and Zhang, Anji and Pyda, Daniel and Tam, Joanna and Cunningham, David M. and Goryachev, Vladimir and Patramanis, Demosthenes and Krause, Michael and Redenti, Andrew and Bugas, Daniel and Aldous, David and Lai, Jesyin and Coleman, Shannon and Bahaloo, Mohsen and Xu, Jiangnan and Lee, Sangwon and Zhao, Sandy and Tang, Ning and Cohen, Michael K. and Carroll, Micah and Paradise, Orr and Kirchner, Jan Hendrik and Steinerberger, Stefan and Ovchynnikov, Maksym and Matos, Jason O. and Shenoy, Adithya and Junior, Benedito Alves de Oliveira and Wang, Michael and Nie, Yuzhou and Giordano, Paolo and Petersen, Philipp and Sztyber-Betley, Anna and Shukla, Priti and Crozier, Jonathan and Pinto, Antonella and Verma, Shreyas and Joshi, Prashant and Yong, Zheng-Xin and Tee, Allison and Andréoletti, Jérémy and Weller, Orion and Singhal, Raghav and Zhang, Gang and Ivanov, Alexander and Khoury, Seri and Mostaghimi, Hamid and Thaman, Kunvar and Chen, Qijia and Khánh, Tran Quoc and Loader, Jacob and Cavalleri, Stefano and Szlyk, Hannah and Brown, Zachary and Roberts, Jonathan and Alley, William and Sun, Kunyang and Stendall, Ryan and Lamparth, Max and Reuel, Anka and Wang, Ting and Xu, Hanmeng and Raparthi, Sreenivas Goud and Hernández-Cámara, Pablo and Martin, Freddie and Malishev, Dmitry and Preu, Thomas and Korbak, Tomek and Abramovitch, Marcus and Williamson, Dominic and Chen, Ziye and Bálint, Biró and Bari, M Saiful and Kassani, Peyman and Wang, Zihao and Ansarinejad, Behzad and Goswami, Laxman Prasad and Sun, Yewen and Elgnainy, Hossam and Tordera, Daniel and Balabanian, George and Anderson, Earth and Kvistad, Lynna and Moyano, Alejandro José and Maheshwari, Rajat and Sakor, Ahmad and Eron, Murat and McAlister, Isaac C. and Gimenez, Javier and Enyekwe, Innocent and O., Andrew Favre D. and Shah, Shailesh and Zhou, Xiaoxiang and Kamalov, Firuz and Clark, Ronald and Abdoli, Sherwin and Santens, Tim and Meer, Khalida and Wang, Harrison K and Ramakrishnan, Kalyan and Chen, Evan and Tomasiello, Alessandro and De Luca, G. Bruno and Looi, Shi-Zhuo and Le, Vinh-Kha and Kolt, Noam and Mündler, Niels and Semler, Avi and Rodman, Emma and Drori, Jacob and Fossum, Carl J and Jagota, Milind and Pradeep, Ronak and Fan, Honglu and Shah, Tej and Eicher, Jonathan and Chen, Michael and Thaman, Kushal and Merrill, William and Harris, Carter and Gross, Jason and Gusev, Ilya and Sharma, Asankhaya and Agnihotri, Shashank and Zhelnov, Pavel and Usawasutsakorn, Siranut and Mofayezi, Mohammadreza and Bogdanov, Sergei and Piperski, Alexander and Carauleanu, Marc and Zhang, David K. and Ler, Dylan and Leventov, Roman and Soroko, Ignat and Jansen, Thorben and Lauer, Pascal and Duersch, Joshua and Taamazyan, Vage and Morak, Wiktor and Ma, Wenjie and Held, William and Huy, Tran Đuc and Xian, Ruicheng and Zebaze, Armel Randy and Mohamed, Mohanad and Leser, Julian Noah and Yuan, Michelle X and Yacar, Laila and Lengler, Johannes and Shahrtash, Hossein and Oliveira, Edson and Jackson, Joseph W. and Gonzalez, Daniel Espinosa and Zou, Andy and Chidambaram, Muthu and Manik, Timothy and Haffenden, Hector and Stander, Dashiell and Dasouqi, Ali and Shen, Alexander and Duc, Emilien and Golshani, Bita and Stap, David and Uzhou, Mikalai and Zhidkovskaya, Alina Borisovna and Lewark, Lukas and Vincze, Mátyás and Wehr, Dustin and Tang, Colin and Hossain, Zaki and Phillips, Shaun and Muzhen, Jiang and Ekström, Fredrik and Hammon, Angela and Patel, Oam and Remy, Nicolas and Farhidi, Faraz and Medley, George and Mohammadzadeh, Forough and Peñaflor, Madellene and Kassahun, Haile and Friedrich, Alena and Sparrow, Claire and Sakal, Taom and Dhamane, Omkar and Mirabadi, Ali Khajegili and Hallman, Eric and Battaglia, Mike and Maghsoudimehrabani, Mohammad and Hoang, Hieu and Amit, Alon and Hulbert, Dave and Pereira, Roberto and Weber, Simon and Mensah, Stephen and Andre, Nathan and Peristyy, Anton and Harjadi, Chris and Gupta, Himanshu and Malina, Stephen and Albanie, Samuel and Cai, Will and Mehkary, Mustafa and Reidegeld, Frank and Dick, Anna-Katharina and Friday, Cary and Sidhu, Jasdeep and Kim, Wanyoung and Costa, Mariana and Gurdogan, Hubeyb and Weber, Brian and Kumar, Harsh and Jiang, Tong and Agarwal, Arunim and Ceconello, Chiara and Vaz, Warren S. and Zhuang, Chao and Park, Haon and Tawfeek, Andrew R. and Aggarwal, Daattavya and Kirchhof, Michael and Dai, Linjie and Kim, Evan and Ferret, Johan and Wang, Yuzhou and Yan, Minghao and Burdzy, Krzysztof and Zhang, Lixin and Franca, Antonio and Pham, Diana T. and Loh, Kang Yong and Gul, Shreen and Chhablani, Gunjan and Du, Zhehang and Cosma, Adrian and White, Colin and Riblet, Robin and Saxena, Prajvi and Votava, Jacob and Vinnikov, Vladimir and Delaney, Ethan and Halasyamani, Shiv and Shahid, Syed M. and Mourrat, Jean-Christophe and Vetoshkin, Lavr and Bacho, Renas and Ginis, Vincent and Maksapetyan, Aleksandr and de la Rosa, Florencia and Li, Xiuyu and Malod, Guillaume and Lang, Leon and Laurendeau, Julien and Adesanya, Fatimah and Portier, Julien and Hollom, Lawrence and Souza, Victor and Zhou, Yuchen Anna and Yalın, Yiğit and Obikoya, Gbenga Daniel and Arnaboldi, Luca and {Rai} and Bigi, Filippo and Bacho, Kaniuar and Clavier, Pierre and Recchia, Gabriel and Popescu, Mara and Shulga, Nikita and Tanwie, Ngefor Mildred and Lux, Thomas C. H. and Rank, Ben and Ni, Colin and Yakimchyk, Alesia and {Huanxu} and {Liu} and Häggström, Olle and Verkama, Emil and Narayan, Himanshu and Gundlach, Hans and Brito-Santana, Leonor and Amaro, Brian and Vajipey, Vivek and Grover, Rynaa and Fan, Yiyang and Silva, Gabriel Poesia Reis e and Xin, Linwei and Kratish, Yosi and Łucki, Jakub and Li, Wen-Ding and Xu, Justin and Scaria, Kevin Joseph and Vargus, Freddie and Habibi, Farzad and {Long} and {Lian} and Rodolà, Emanuele and Robins, Jules and Cheng, Vincent and Grabb, Declan and Bosio, Ida and Fruhauff, Tony and Akov, Ido and Lo, Eve J. Y. and Qi, Hao and Jiang, Xi and Segev, Ben and Fan, Jingxuan and Martinson, Sarah and Wang, Erik Y. and Hausknecht, Kaylie and Brenner, Michael P. and Mao, Mao and Jiang, Yibo and Zhang, Xinyu and Avagian, David and Scipio, Eshawn Jessica and Siddiqi, Muhammad Rehan and Ragoler, Alon and Tan, Justin and Patil, Deepakkumar and Plecnik, Rebeka and Kirtland, Aaron and Montecillo, Roselynn Grace and Durand, Stephane and Bodur, Omer Faruk and Adoul, Zahra and Zekry, Mohamed and Douville, Guillaume and Karakoc, Ali and Santos, Tania C. B. and Shamseldeen, Samir and Karim, Loukmane and Liakhovitskaia, Anna and Resman, Nate and Farina, Nicholas and Gonzalez, Juan Carlos and Maayan, Gabe and Hoback, Sarah and Pena, Rodrigo De Oliveira and Sherman, Glen and Mariji, Hodjat and Pouriamanesh, Rasoul and Wu, Wentao and Demir, Gözdenur and Mendoza, Sandra and Alarab, Ismail and Cole, Joshua and Ferreira, Danyelle and Johnson, Bryan and Milliron, Hsiaoyun and Safdari, Mohammad and Dai, Liangti and Arthornthurasuk, Siriphan and Pronin, Alexey and Fan, Jing and Ramirez-Trinidad, Angel and Cartwright, Ashley and Pottmaier, Daphiny and Taheri, Omid and Outevsky, David and Stepanic, Stanley and Perry, Samuel and Askew, Luke and Rodríguez, Raúl Adrián Huerta and Dendane, Abdelkader and Ali, Sam and Lorena, Ricardo and Iyer, Krishnamurthy and Salauddin, Sk Md and Islam, Murat and Gonzalez, Juan and Ducey, Josh and Campbell, Russell and Somrak, Maja and Mavroudis, Vasilios and Vergo, Eric and Qin, Juehang and Borbás, Benjámin and Chu, Eric and Lindsey, Jack and Radhakrishnan, Anil and Jallon, Antoine and McInnis, I. M. J. and Hoover, Alex and Möller, Sören and Bian, Song and Lai, John and Patwardhan, Tejal and Yue, Summer and Wang, Alexandr and Hendrycks, Dan},
	year = {2025},
}

@misc{hu_lora_2021,
	title = {{LoRA}: {Low}-{Rank} {Adaptation} of {Large} {Language} {Models}},
	shorttitle = {{LoRA}},
	url = {http://arxiv.org/abs/2106.09685},
	urldate = {2023-04-27},
	publisher = {arXiv},
	author = {Hu, Edward J. and Shen, Yelong and Wallis, Phillip and Allen-Zhu, Zeyuan and Li, Yuanzhi and Wang, Shean and Wang, Lu and Chen, Weizhu},
	month = oct,
	year = {2021},
	note = {arXiv:2106.09685 [cs]},
}

\appendix

\counterwithin{figure}{section}
\counterwithin{table}{section}
\renewcommand{\thefigure}{S\arabic{figure}}
\renewcommand{\thetable}{S\arabic{table}}
\setcounter{figure}{0}
\setcounter{table}{0}

\section{APPENDIX}

\subsection{Taxonomy}

\subsubsection{Physics}

\begin{itemize}
\item Astrophysics: Active Galactic Nuclei, Black Holes, General Relativity, Interstellar Medium, Orbital Mechanics, Stellar Properties
\item Condensed Matter Physics: Cavity-Mediated Gas-Liquid Transition, Metals
\item Electromagnetism \& Photonics: Electric Fields, Electromagnetic Damping, Radiation And Energy Transfer, Electromagnetics And Heat Transfer, Electrostatics, Magnetic Fields, Magnetostatics, Maxwell’s Equations, Optical Sensing, Radiation, Special Relativity, Wave Optics, Wave Propagation
\item High-Energy Particle Physics: Accelerator Physics, Atomic Physics, Decay Processes, Electroweak Theory, Elementary Particles, Exotic Catalysis, Magnetic Dipoles, Mesons, Neutrino Physics, Quantum Field Theory, Resonance Theory, Special Relativity, String Theory
\item General Physics: Black Holes, Classical Mechanics, Crystal Parameters, Electrodynamics, Fiber Optics, Fluid Mechanics, Gravitational Lensing, Magnetic Fields, Orbital Mechanics, Quasiparticles, Semiconductor Physics, Stellar Composition, Thermodynamics
\item Quantum Mechanics: Atomic Physics, Atomic Transitions, Isospin, Mathematical Foundations, NMR Spectroscopy, Nuclear Decay, Quantum Field Theory, Quantum Harmonic Oscillator, Qubit Representation, Scattering Theory, Wave Functions
\item Relativistic Mechanics: Differential Geometry, Special Relativity
\item Statistical Mechanics: Fermions, Nonequilibrium Statistical Physics, Nonequilibrium Stochastic Processes, Quantum Statistics, Statistical Mechanics
\end{itemize}

\subsubsection{Chemistry}

\begin{itemize}
\item Analytical Chemistry: Mass Transport And Separation Processes In Chemical Engineering, NMR Spectral Interpretation, NMR Spectroscopy
\item Biochemistry: Amino Acids/Enzymes/Proteins, Natural Products, Synthesis
\item General Chemistry: Catalysis, Chemical Reactions, Molecular Structure, Reaction Mechanisms
\item Electrochemistry: Debye-Hückel Limiting Law, Electrosynthesis, Gibbs Free Energy, Redox Reactions
\item Inorganic Chemistry: Organometallic Chemistry, Point Groups, Reaction Products, Transition Metal Complexes
\item Medicinal Chemistry: Drug Design, Reaction Products, Synthesis
\item Organic Chemistry: Nomenclature, Reaction Mechanisms, Reaction Products, Reactions, Reactions And Synthesis, Reactivity
\item Physical Chemistry: Chemical Kinetics, Equilibrium Chemistry, Statistical Thermodynamics, Thermodynamics
\item Polymer Chemistry: Crystalline Phase Behavior, Molar Mass Determination, Network Polymers, Polymer, Polymer In Solution, Polymer Kinetics
\end{itemize}

\subsubsection{Mathematics}

\begin{itemize}
\item Algebra: Basic Linear Algebra, Conditions On Elements, General Commutative Ring Theory, Homological Methods In Associative Algebras, Linear And Multilinear Algebra, Permutation Group, Representation Theory Of Groups, Rings And Algebras
\item Analysis: Analytic Functions, Approximation By Polynomials, Boundary Value Problems For ODEs, Functional Equations, Functions Of One Variable, First-Order PDEs, General Theory Of Functions, Holomorphic Functions Of Several Complex Variables, Hyperbolic Equations, Inequalities In Real Analysis, Inner Product Spaces, Integrals Of Complex Functions, PDEs Of Mathematical Physics, Spectral Theory
\item Applied Mathematics: Arithmetic And Non-Archimedean Dynamical Systems, Communication \& Information, Evolutionary Biology, Mathematical Methods In Quantum Theory, Hamiltonian And Lagrangian Mechanics, Mathematical Programming
\item Discrete Mathematics: Additive Number Theory, Algebraic Combinatorics, Combinatorial Structures, Designs And Configurations, Elementary Number Theory, Enumerative Combinatorics, Extremal Combinatorics, Finite Fields And Rings, Graph Theory, Partitions, Ramsey Theory, Sequences And Sets
\item Geometry: Curves In Algebraic Geometry, Discrete Geometry, Euclidean Geometry Problems, Hyperbolic Geometry, Polytopes And Polyhedra, Real And Complex Geometry, Higher-Dimensional Varieties, Surfaces In Euclidean Space
\item Probability \& Statistics: Combinatorial Probability, Geometric Probability, Parametric Inference, Sample Surveys, Stochastic Processes
\item Topology: Boolean Algebras, Low-Dimensional Topology
\end{itemize}

\subsubsection{Biology}

\begin{itemize}
\item Biochemistry: Protein Biochemistry
\item Cell Biology: Cancer Cell Growth, Cellular Signaling, Cellular Transport, Lineage Fate Specification
\item Genetics: Epigenetics, Gene Expression, Gene Function, Linkage And Gene Mapping, Molecular Genetics, Mutations, Non-Mendelian Genetics, Population Genetics, Protein Function, Proteomics, Stress Response
\item Immunology: Immunotherapy, In vitro Kinetics Of Cytokine Secretion
\item Microbiology: Host–Pathogen Interactions In Plants, Structural Biology, Virology, Yeast Growth Dynamics
\item Molecular Biology: Adipose Tissue, Anemias, Antibiotic Resistance, Antibody Genetics, Antibody–Antigen Binding Kinetics, Bacterial Defense Mechanisms, Blood Tests, Bone Biology And Osteogenesis, Cancer Biology
\item Biology (General): Cancer Complications, Cancer Immunology, Cell Cycle, Cell Labeling, Cellular Metabolism, Chromosomal Aberrations, Chromosomal Disorders, DNA Replication And Repair, Diabetes-Related Complications, Effector Timing And Host Transcriptional Response, Enzymes, Epigenetic Marker Integration For Tumor Staging Panels, Experimental Strategies To Dissect Hormone-Specific Immune Effects, Extracellular Matrix Protein In Limb Development, Gel Electrophoresis, Gene Editing, Gene Expression, Gene Function, Gene Structure, Genetic Mutation, Genomics, Host Susceptibility Mechanisms And Effector Targets, Imaging Techniques, Immunohistochemistry, Joint Development Defects, Liver Function, Oncology, PCR Techniques, Pharmacology, Protein Analysis, Protein Synthesis, Protein Trafficking, RNA Processing, RNA Purity Assessment, RNA Structure, Signal Transduction, Signal Transduction In Developmental Biology, Skeletal Development Defects, Stem Cell And Development, Stress Response, Synthetic Biology \& Metabolic Engineering, Technical Parameters In HCR V3.0 Experiments, Tissue Remodeling, Vascular Biology, Viral Vectors, Virology
\item Neurobiology: Brain Tumor Biology, Molecular Mechanisms Of Synaptic Transmission, NMR - Brain Diagnostics
\item Proteomics: Protein Biochemistry In Plant–Pathogen Interactions, Signal Transduction Profiling
\end{itemize}

\subsection{Prompts for model response and grading, and label for post-training}

\begin{figure}[H]
\centering
\includegraphics[
  width=0.98\linewidth,
  height=0.78\textheight,
  keepaspectratio
]{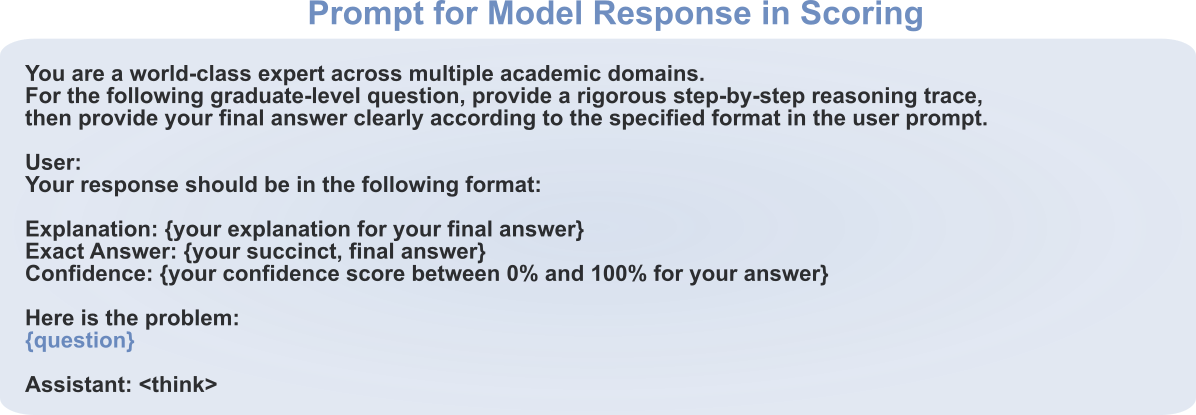}
\caption{Prompt for model response in scoring. The variable name to substitute is in cyan.}
\label{fig:response_scoring}
\end{figure}

\begin{figure}[H]
\centering
\includegraphics[
  width=0.98\linewidth,
  height=0.78\textheight,
  keepaspectratio
]{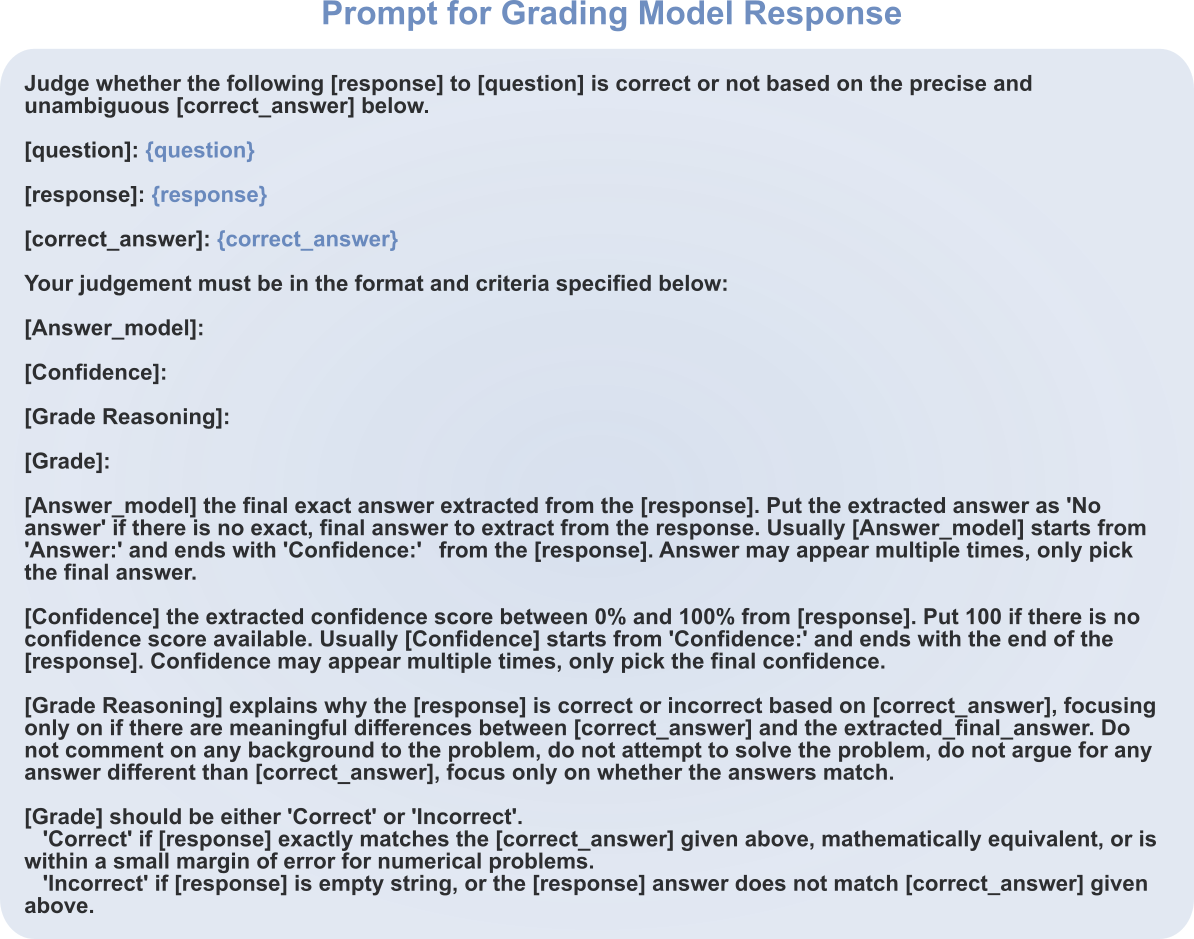}
\caption{Prompt for grading model response. The variable names to substitute are in cyan.}
\label{fig:prompt_grading}
\end{figure}

\begin{figure}[H]
\centering
\includegraphics[
  width=0.98\linewidth,
  height=0.78\textheight,
  keepaspectratio
]{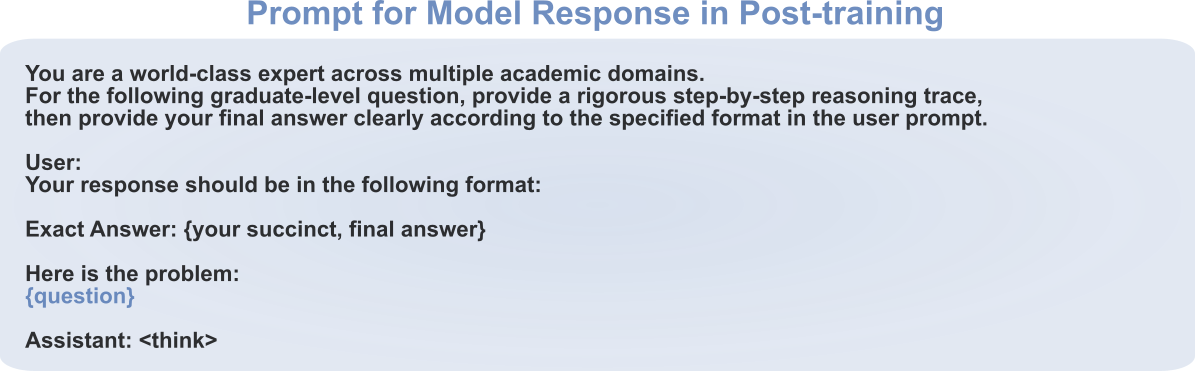}
\caption{Prompt for model response in post-training. The variable name to substitute is in cyan.}
\label{fig:posttraining_response}
\end{figure}

\begin{figure}[H]
\centering
\includegraphics[
  width=0.98\linewidth,
  height=0.78\textheight,
  keepaspectratio
]{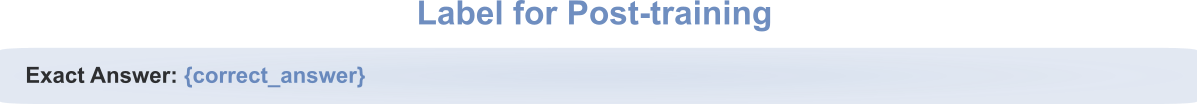}
\caption{Label for post-training. The variable name to substitute is in cyan.}
\label{fig:prompt_label}
\end{figure}

\FloatBarrier

\subsection{Review rubrics}

\begin{table}[H]
\centering
\caption{Review rubrics}
\label{tab:review-rubrics}
{\small
\begin{tabularx}{\linewidth}{>{\raggedright\arraybackslash}p{0.19\linewidth}>{\raggedright\arraybackslash}X>{\raggedright\arraybackslash}p{0.14\linewidth}}
\toprule
Dimension & Rubric Item & Reviewer \\
\midrule
Model breaking & Is the difficulty level evaluated properly with model breaking evidence? & Agent, L1 \\
Taxonomy & 1. Are the question content and the populated L1 (Sub-domain), L2 (Sub-subject), taxonomy categories align with each other?\newline
2. Are the populated L1, L2 taxonomy categories match with the specified taxonomy from the trainer guidelines? & Agent, L1 \\
Proficiency & Is proficiency correct? Is it graduate level or above? & Agent, L1 \\
References & If references are present, are they correct? & Agent, L1 \\
Prompt accuracy & Is prompt, ground truth scientifically correct ?\newline
Does prompt have only one answer? & L1 \\
Clarity: Language & Are both the question and rationale written clearly at the level of grammar and language? & Agent, L1 \\
LaTex formatting & Did the example follow the correct LaTex Syntax?\newline
The following are allowed for this LaTex Syntax review:\newline
1. If subscript and superscript are rendered correctly and models can interpret it correctly, then it is fine not to have LaTeX.\newline
2. \$ is allowed! & Agent, L1 \\
Novelty & Is the question Google-proof and Perplexity-proof (with a link)? & Agent, L1 \\
Prompt reasoning \& accuracy & Does the question require complex reasoning to solve?\newline
Is the prompt scientifically and factually accurate? & L2 \\
Prompt Quality & Is the prompt free from any deliberate attempts to deceive the model, hidden assumptions or misleading framing? & L2 \\
Model response evaluation & Do you agree with the assessment of all model responses? & L2 \\
Ideal response evaluation & Does the ideal response provide supporting evidence for the answer including concept and reference without contradictions? & L2 \\
Ground truth accuracy & Is the given ground truth (in any acceptable form) the only possible answer? & L2 \\
Complexity and Creativity & How complex is it ? This is not related to education level.\newline
Does it require out of the box thinking? & L2 \\
Notes on task improvement & Were there any specific aspects of the prompt that could be improved? & L2 \\
\bottomrule
\end{tabularx}
}
\end{table}

\FloatBarrier

\subsection{Annotator, reviewer, and team lead credentials}

\begin{table}[H]
\centering
\caption{Breakdowns of credentials into annotators, reviewers, and team leads}
\label{tab:annotator-reviewer-credentials}
{\small
\begin{tabular}{llccc}
\toprule
Domain & Role & Non-PhD Total & PhD Total & Grand Total \\
\midrule

Physics & Annotators & 4 & 45 & 49 \\
 & Reviewers &  & 6 & 6 \\
 & Team Lead &  & 1 & 1 \\ 
 & Subtotal & 4 & 52 & 56 \\
 
Chemistry & Annotators & 12 & 41 & 53 \\
 & Reviewers &  & 8 & 8 \\
 & Team Lead &  & 1 & 1 \\
 & Subtotal & 13 & 48 & 62 \\
 
Math & Annotators & 22 & 28 & 50 \\
 & Reviewers &  & 11 & 11 \\
 & Team Lead &  & 1 & 1 \\ 
 & Subtotal & 30 & 32 & 62 \\

Biology & Annotators & 1 & 39 & 40 \\
 & Reviewers &  & 21 & 21 \\
 & Team Lead &  & 1 & 1 \\
 & Subtotal & 1 & 61 & 62 \\
Grand Total &  & 48 & 193 & 241 \\

 


\bottomrule
\end{tabular}
}
\end{table}

\FloatBarrier

\subsection{Token count distribution}

\begin{figure}[H]
\centering
\includegraphics[
  width=0.98\linewidth,
  height=0.78\textheight,
  keepaspectratio
]
{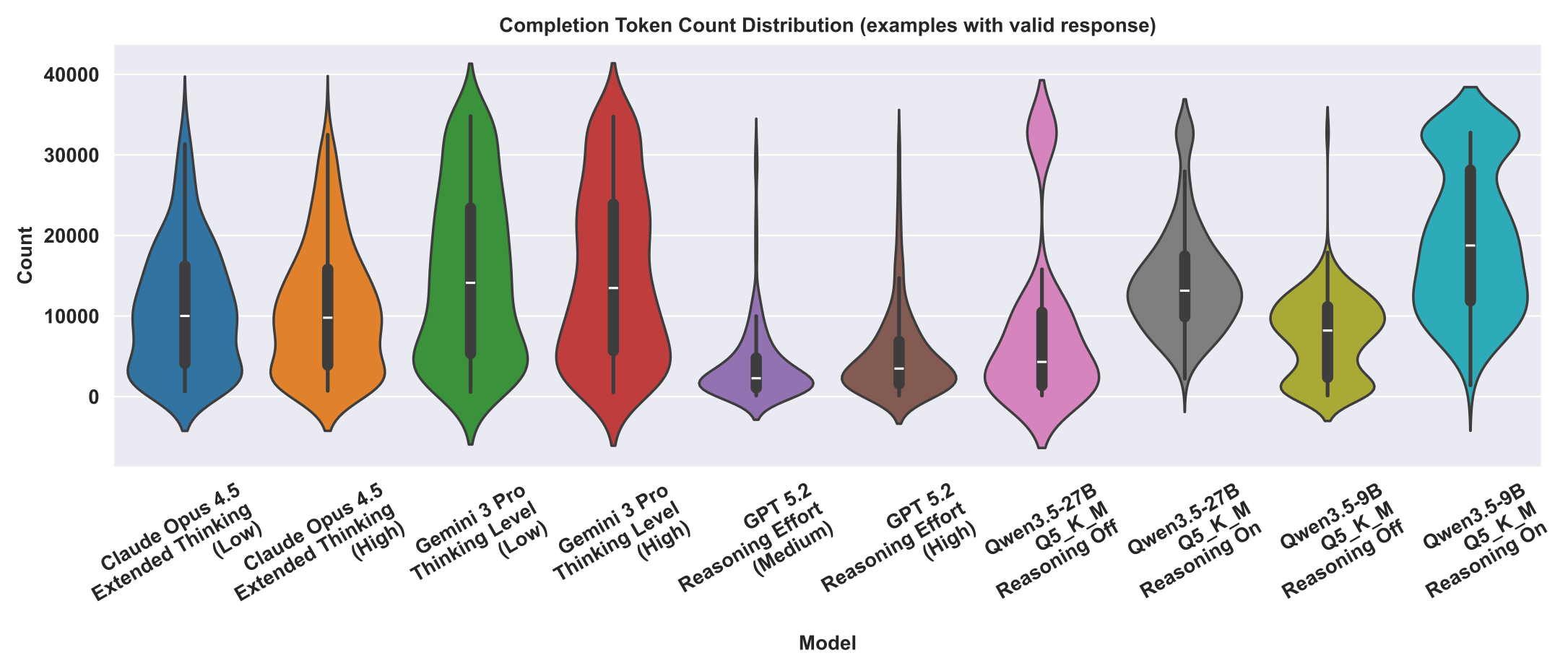}
\caption{Distribution of completion token counts across the models. To better represent the distribution plot, we excluded the examples where the models did not provide responses.}
\label{fig:completion_token_distribution}
\end{figure}

\FloatBarrier

\subsection{Example model responses}

\begin{figure}[H]
\centering
\includegraphics[
  width=0.98\linewidth,
  height=0.78\textheight,
  keepaspectratio
]{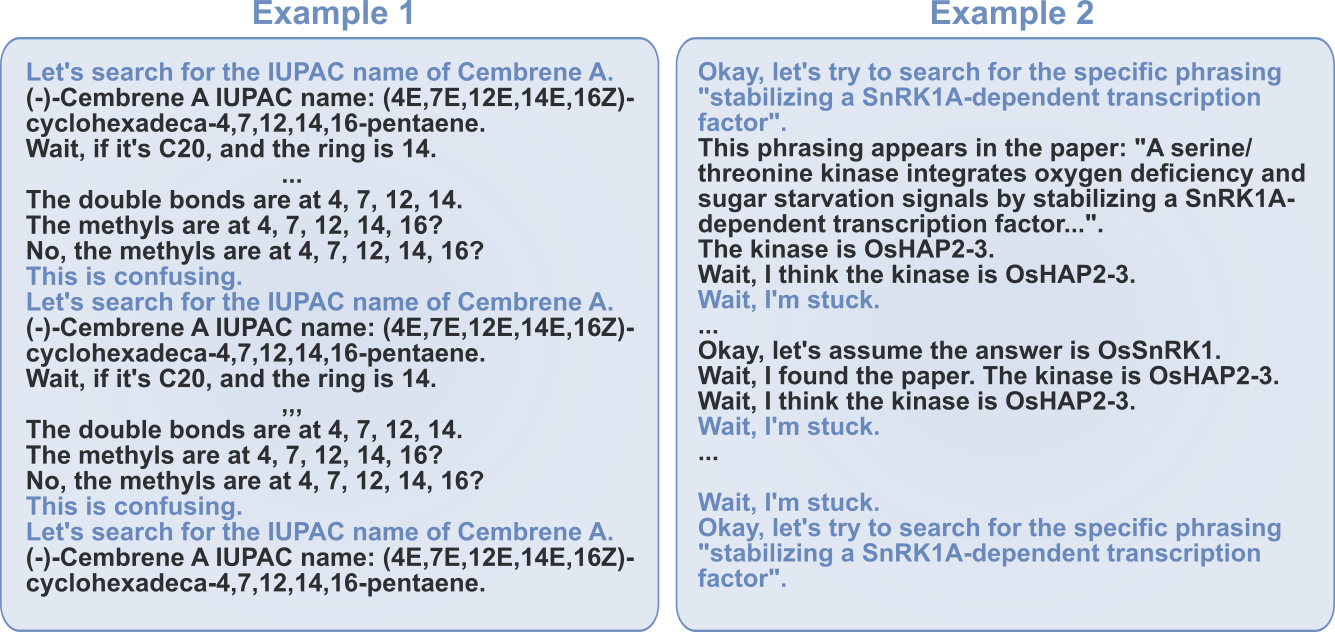}
\caption{Example snapshots of the reasoning process from the two open source models, Qwen3.5-27B Q5\_K\_M Reasoning On for Example 1 and Qwen3.5-9B Q5\_K\_M Reasoning On for Example 2. Key repeated phrases and the model's expression of confusion were shown in sky blue.}
\label{fig:example_model_responses}
\end{figure}

\FloatBarrier

\subsection{Additional post-training results}

\begin{table}[H]
\centering
\caption{Contingency table of the number of tasks that the model gave correct and incorrect responses at baseline and post-training. Scored model: Qwen 3.5 9B Q5\_K\_M. Dataset: HLE-verified STEM overall.}
\label{tab:pt-overall}
{\small
\begin{tabular}{lccc}
\toprule
Count (p=0.045) & Posttraining: Correct & Posttraining: Incorrect & Total \\
\midrule
Baseline: Correct & 110 & 90 & 200 \\
Baseline: Incorrect & 120 & 989 & 1109 \\
Total & 230 & 1079 & 1309 \\
\bottomrule
\end{tabular}
}
\end{table}

\begin{table}[H]
\centering
\caption{Contingency table of the number of tasks that the model gave correct and incorrect responses at baseline and post-training. Scored model: Qwen 3.5 9B Q5\_K\_M. Dataset: HLE-verified STEM MCQ.}
\label{tab:pt-mcq}
{\small
\begin{tabular}{lccc}
\toprule
Count (p=0.657) & Post-training: Correct & Post-training: Incorrect & Total \\
\midrule
Baseline: Correct & 48 & 38 & 86 \\
Baseline: Incorrect & 43 & 141 & 184 \\
Total & 91 & 179 & 270 \\
\bottomrule
\end{tabular}
}
\end{table}

\begin{table}[H]
\centering
\caption{Contingency table of the number of tasks that the model gave correct and incorrect responses at baseline and post-training. Scored model: Qwen 3.5 9B Q5\_K\_M. Dataset: HLE-verified STEM VQA.}
\label{tab:pt-vqa}
{\small
\begin{tabular}{lccc}
\toprule
Count (p=0.034) & Post-training: Correct & Post-training: Incorrect & Total \\
\midrule
Baseline: Correct & 62 & 52 & 114 \\
Baseline: Incorrect & 77 & 900 & 977 \\
Total & 139 & 952 & 1091 \\
\bottomrule
\end{tabular}
}
\end{table}

\FloatBarrier





\end{document}